\documentclass{article}
\usepackage{iclr2027_conference,times}

\usepackage{amsmath,amsfonts,bm}

\def\eqref#1{equation~\ref{#1}}

\def\1{\bm{1}}

\DeclareMathAlphabet{\mathsfit}{\encodingdefault}{\sfdefault}{m}{sl}
\SetMathAlphabet{\mathsfit}{bold}{\encodingdefault}{\sfdefault}{bx}{n}

\usepackage{hyperref}
\hypersetup{hidelinks}
\usepackage{enumitem}
\usepackage{url}
\usepackage{booktabs}
\usepackage{amsmath}
\usepackage{amssymb}
\usepackage{graphicx}
\usepackage{array}
\usepackage{tabularx}
\usepackage{algorithm}
\usepackage{algpseudocode}
\usepackage{capt-of}
\usepackage{float}
\floatstyle{plain}
\newfloat{algorithmpair}{tbp}{lop}

\usepackage{multirow}
\usepackage[table]{xcolor}
\definecolor{carow}{RGB}{230,242,255} 
\definecolor{ablationblue}{HTML}{245B78}
\definecolor{ablationgray}{HTML}{707070}

\usepackage{textcomp}

\definecolor{efredefaultbg}{HTML}{EAF2F8}
\definecolor{baselinebg}{HTML}{F5F0E6}
\definecolor{efreteacherbg}{HTML}{C6DCEF}
\definecolor{efrebestbg}{HTML}{F8EDCB}
\definecolor{efreoraclegray}{HTML}{777777}

\title{From a Prompt to Repertoires: Evolving Functional REpertoires Enable LLM Continual Learning}

\author{%
\textbf{Fengyuan Liu}$^{1}$,
\textbf{Yue Wang}$^{2,3}$,
\textbf{Hangxi Guo}$^{1}$,
\textbf{Fengyuan Liu}$^{4}$,
\textbf{Chenxu Wu}$^{3}$,\\
\textbf{Yanguang Liu}$^{5}$,
\textbf{Mengnan Du}$^{1,\dagger}$
\\
$^{1}$The Chinese University of Hong Kong, Shenzhen
$^{2}$Shanghai AI Laboratory \\
$^{3}$University of Science and Technology of China
$^{4}$Independent Researcher \\
$^{5}$New Jersey Institute of Technology \\
\texttt{liuferry708@gmail.com, mengnandu@cuhk.edu.cn} \\
$^{\dagger}$Corresponding author
}

\iclrfinalcopy

\begin{document}

\maketitle

\lhead{Preprint}

\begin{abstract}
Continual learning remains challenging for large language models, which must enable models to acquire new skills and knowledge without degrading existing capabilities.
Existing approaches typically address this challenge by carefully
designing how model parameters are updated.
In contrast, prompt optimization avoids costly parameter updates
while achieving competitive or even superior performance to
reinforcement learning methods such as GRPO on individual
knowledge-intensive and reasoning tasks.
This raises a natural question:
\textit{Can prompt optimization, as an efficient adaptation
approach, be directly applied to continual learning?}
Our analysis shows that, under sequential task adaptation, it suffers
from catastrophic forgetting, while optimized prompts accumulate rules
that overfit to local task distributions.
To address these limitations, we propose
\emph{Evolving Functional REpertoires} (EFRE), which replaces
a single prompt with a repertoire of functions that evolves
as new tasks arrive: compatible updates refine existing
functions, while conflicting updates trigger the emergence
of new ones.
On a three-task continual-learning stream, EFRE achieves a final
average performance 7.50 percentage points higher than GRPO.
Moreover, after adaptation to the Bio task, its performance on
FinQA decreases by only 1.56 percentage points, compared with
25.10 percentage points for the base prompt optimization method.
We further instantiate EFRE in a minimal agent system and observe
consistent improvements across different backbone models.
Overall, these results demonstrate EFRE's strong performance
in continual learning for large language models and highlight its
substantial potential for continual learning in advanced agent
systems.
\end{abstract}

\section{Introduction}

Continual learning is becoming increasingly important for large language models (LLMs), which need to continually acquire new knowledge and skills while preserving what they have already learned~\citep{harrington2026continual,asawa2026clbench}.
Most existing approaches address this challenge by controlling how model parameters are updated, for example through low-rank adaptation (LoRA)~\citep{hu2021lora}, replay-based training~\citep{lopezpaz2017gem}, or reinforcement-learning-based optimization such as GRPO~\citep{shao2024deepseekmath} rely on reliable rewards~\citep{tian2026autorubric,liu2026dynacf}.
Although effective, these methods incur substantial training
overhead. Enabling efficient continual learning 
remains an open challenge.

Recently, prompt optimization~\citep{pryzant2023apo,yuksekgonul2024textgrad,agrawal2025gepa,zhang2025ace}
has emerged as an efficient approach to adapting LLMs
to downstream tasks without parameter updates.
These methods have achieved strong performance on
knowledge-intensive and reasoning tasks, in some cases
outperforming reinforcement learning methods such as GRPO. This naturally raises the question: \textit{Can prompt optimization, as a powerful and efficient optimization method, be directly applied to continual learning?} Our analysis and prior studies~\citep{harrington2026continual} suggest otherwise:
under sequential multi-task learning, repeated prompt updates
overwrite previously learned knowledge and accumulate
case-specific rules rather than generalizable principles,
limiting their effectiveness in continual learning.

Motivated by these limitations, we introduce
\emph{Evolving Functional REpertoires} (\textbf{EFRE}),
a continual learning method that evolves both the
capabilities and composition of a functional repertoire
as new learning tasks arrive.
Each function is represented by an optimized prompt.
Historical consistency checks guide this evolution:
compatible updates refine existing functions and enable
reuse across tasks, while updates that cannot be absorbed
without compromising historical capabilities trigger
the emergence of new functions.
Through this process, functional specialization and reuse
arise automatically, reducing interference between new
and previously learned capabilities.

We evaluate EFRE on the ToolUse~\citep{tang2023toolalpaca} $\rightarrow$ FinQA~\citep{chen2021finqa} $\rightarrow$ SciKE-Bio~\citep{feng2024sciknoweval}
continual-learning sequence and several single-task benchmarks. With Qwen3-8B~\citep{qwen2025eightbcard}, EFRE achieves a final average score of 58.75, outperforming GRPO~\citep{shao2024deepseekmath} and GEPA~\citep{agrawal2025gepa}, while limiting forgetting on previously learned tasks after Bio to 1.57 points, versus 8.32 for GRPO and 11.82 for GEPA. EFRE also consistently improves over prompt-optimization and GRPO baselines on single-task benchmarks. In a minimal agent system spanning six continual-learning tasks, EFRE similarly outperforms ICL~\citep{brown2020fewshot} and GEPA: it attains average scores of 18.0 with Qwen3-8B (vs.\ $-3.5$ and 9.5) and 36.3 with GPT-5.6-Luna~\citep{openai2026luna} (vs.\ 22.2 and 29.5). Together, these results demonstrate EFRE's effectiveness for both standard and agent continual learning.

Our key contributions are threefold:
\begin{itemize}[leftmargin=10pt, topsep=-2pt, itemsep=1pt, partopsep=1pt, parsep=1pt]
    \item We identify two key limitations of prompt optimization
    in continual learning: updates overwrite previously
    learned knowledge, causing forgetting, and prompts
    accumulate case-specific rules rather than generalizable
    principles.
    \item We introduce EFRE, a non-parametric continual learning
    method whose functional repertoire evolves with incoming
    tasks through the refinement and reuse of existing
    functions and the emergence of new ones, guided by
    historical consistency checks.
    \item We empirically demonstrate that EFRE improves both
    continual learning and single-task optimization,
    outperforming existing prompt optimization methods
    and GRPO in overall performance.
    Further gains in agent continual-learning settings
    highlight its potential for continual learning in real-world intelligent systems.
\end{itemize}

\section{Related Work}

\textbf{Continual Learning.}
Continual learning is a learning paradigm that requires a model
to acquire new capabilities while preserving previously learned
knowledge and behaviors. Existing approaches mainly address the stability--plasticity
trade-off through replay~\citep{lopezpaz2017gem}, regularization~\citep{kirkpatrick2017overcoming}, parameter isolation~\citep{rusu2016progressive},
parameter-efficient tuning~\citep{hu2021lora}, and reinforcement-learning-based optimization~\citep{shao2024deepseekmath}. Recent work has further extended continual learning to LLMs and
language agents, where models must continually adapt to new tasks,
knowledge, and interaction environments~\citep{shenfeld2026selfdistillation,asawa2026clbench}. EFRE studies this problem from a non-parametric perspective, maintaining and evolving external functional contexts rather than
continually updating model parameters.

\textbf{Non-Parametric Optimization.}
Non-parametric optimization is a class of lightweight optimization methods
that improves model behavior without updating model parameters,
with prompt optimization being a representative approach~\citep{pryzant2023apo,yuksekgonul2024textgrad}.
Existing methods optimize prompts, contexts, memories, or other
external textual states using task feedback, reflection, or
iterative search~\citep{shinn2023reflexion,agrawal2025gepa,zhang2025ace}. Recent prompt optimization methods have demonstrated strong
performance across reasoning, knowledge-intensive, and agent tasks,
showing that textual states can serve as effective optimization
targets~\citep{agrawal2025gepa,zhang2025ace}. Building on this, EFRE organizes the evolving
non-parametric learning state into a repertoire of functional
contexts, enabling prompt optimization to support continual learning
across sequential tasks.

\begin{figure}[t]
    \centering
    \includegraphics[width=\linewidth]{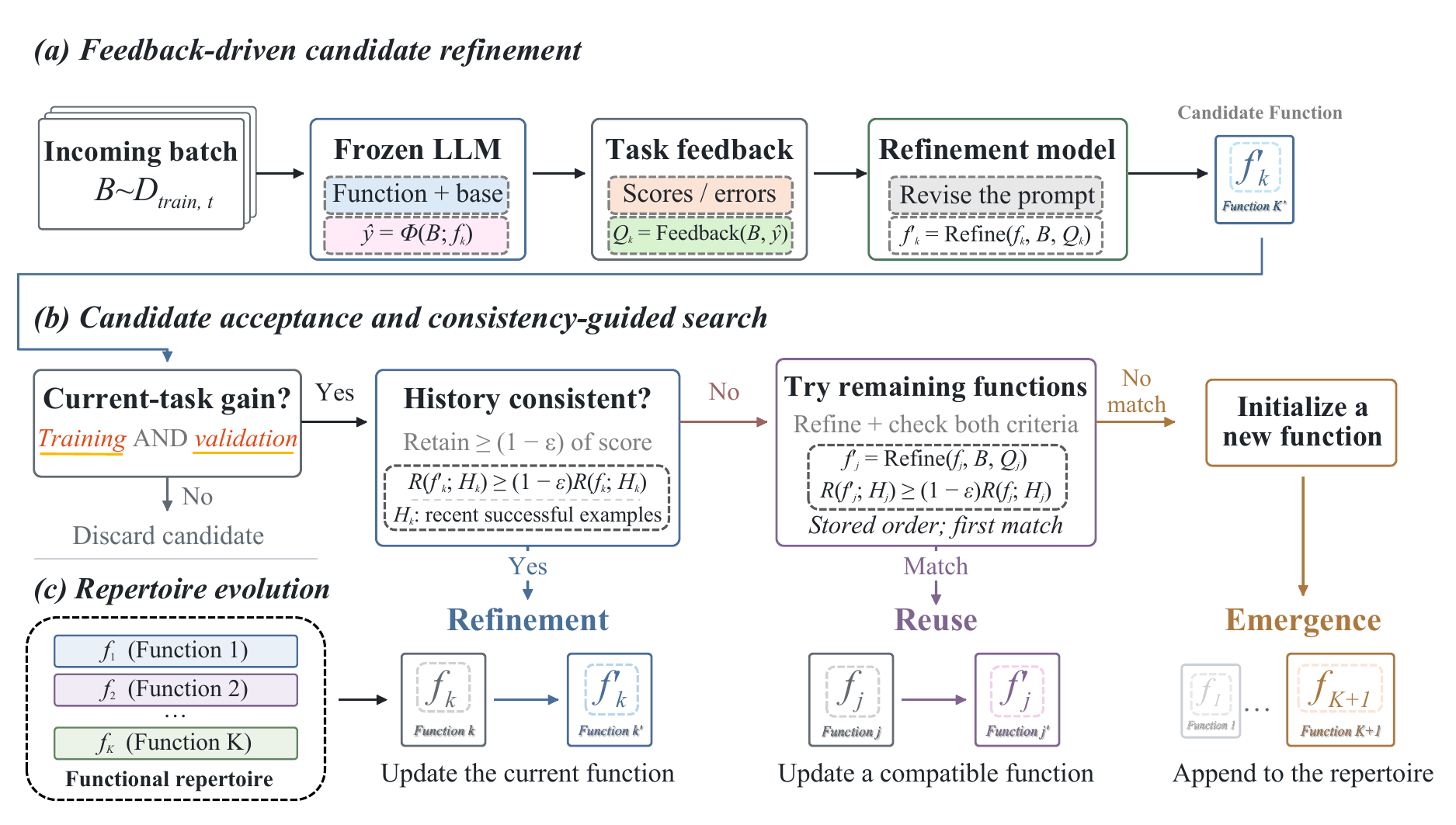}
    \vspace{-15pt}
    \caption{
    Overview of Evolving Functional REpertoires (EFRE).
    Each function is represented by a prompt and evolves through
    feedback-driven refinement on incoming data.
    Candidates that improve current performance undergo a historical
    consistency check: compatible updates refine the current function,
    while incompatible updates trigger a sequential search for another
    function that can absorb them.
    If no compatible update is found, a new function emerges.
    The repertoire thus evolves through functional refinement, reuse,
    and emergence, enabling continual adaptation while preserving
    previously learned capabilities.
    }
    \label{fig:overview}
\end{figure}

\section{Methodology}
\label{sec:method}

In this section, we introduce \emph{Evolving Functional
REpertoires} (EFRE) .
We first formulate continual learning with frozen LLMs.
We then describe prompt optimization for continual
learning and discuss its limitations.
Finally, we present EFRE, which uses historical consistency
checks to guide the refinement of existing functions
and the emergence of new ones as learning tasks arrive (Figure~\ref{fig:overview}).

\subsection{Problem Statement}
\label{sec:problem}

We study continual learning for a pretrained language
model $\Phi$ whose parameters remain fixed.
At stage $t$, the learner receives a training batch
$D_{\mathrm{train},t}$ and updates its external textual
state through
$S_t=\mathcal{U}(S_{t-1},D_{\mathrm{train},t})$,
where $\mathcal{U}$ is the update procedure.
The state may contain prompts, skills, or other textual
information that guides the model's behavior.
The objective is to acquire new capabilities while
preserving those learned at earlier stages.

For an evaluation set $D_{\mathrm{eval}}$, we define
performance as:
\begin{equation}
    R(S;D_{\mathrm{eval}})
    =
    \mathrm{Eval}\!\left(\Phi(\cdot;S),D_{\mathrm{eval}}\right),
    \label{eq:performance}
\end{equation}
where $\mathrm{Eval}$ scores the model conditioned on
$S$, with higher values indicating better performance.
We use the notation $R$ when scoring training,
validation, and history data.
For continual-learning evaluation,
$D_{\mathrm{eval},i}$ denotes the held-out evaluation
set for stage $i$ and is never used to update $S$.

\textbf{Plasticity and Stability.}
We characterize each update through two classical properties
of continual learning.
\emph{Plasticity} measures its performance gain on
the current task:
\begin{equation}
    \mathcal{J}_{\mathrm{plas}}^{(t)}
    =
    R(S_t;D_{\mathrm{eval},t})
    -
    R(S_{t-1};D_{\mathrm{eval},t}).
    \label{eq:plasticity}
\end{equation}
A larger value indicates more effective learning
from the incoming data.
\emph{Stability} measures the average performance
change on previously encountered tasks:
\begin{equation}
    \mathcal{J}_{\mathrm{stab}}^{(t)}
    =
    \frac{1}{t-1}
    \sum_{i=1}^{t-1}
    \left[
        R(S_t;D_{\mathrm{eval},i})
        -
        R(S_{t-1};D_{\mathrm{eval},i})
    \right],
    \; t>1.
    \label{eq:stability}
\end{equation}
Negative values indicate average forgetting, zero
indicates no net change in historical performance,
and positive values indicate beneficial backward
transfer~\citep{lopezpaz2017gem}.
Effective continual learning should achieve positive
plasticity while limiting degradation on previous tasks.
\subsection{Prompt Optimization for Continual Learning}
\label{sec:prompt-optimization}

In a continual learning setting, prompt optimization implements
$S_t=\mathcal{U}(S_{t-1},D_{\mathrm{train},t})$ by
rewriting the prompt represented by $S_{t-1}$.
At stage $t$, the frozen model runs on
$D_{\mathrm{train},t}$ using the current prompt.
The task environment returns feedback $\mathcal{Q}_t$,
including performance scores and, when available,
textual feedback such as test results, error messages,
or explanations of incorrect answers.
An external refinement model $M_{\mathrm{refine}}$
uses this feedback and the training data to revise
$S_{t-1}$:
\begin{equation}
    S_t'
    =
    M_{\mathrm{refine}}
    (S_{t-1},D_{\mathrm{train},t},\mathcal{Q}_t).
    \label{eq:prompt-refinement}
\end{equation}
The candidate replaces the current prompt only if
it improves performance on both the training batch
and a separate validation set:
\begin{equation}
    \begin{aligned}
        R(S_t';D_{\mathrm{train},t})
        &>R(S_{t-1};D_{\mathrm{train},t}),\\
        R(S_t';D_{\mathrm{val},t})
        &>R(S_{t-1};D_{\mathrm{val},t}).
    \end{aligned}
    \label{eq:prompt-update}
\end{equation}
Otherwise, $S_t=S_{t-1}$.
The validation set $D_{\mathrm{val},t}$ is used for
update selection and is distinct from the held-out
evaluation set $D_{\mathrm{eval},t}$.

Our analysis identifies limitations in both stability
and plasticity.
The acceptance rule checks current performance but
does not explicitly assess retention on previous tasks.
Repeated rewriting can therefore overwrite instructions
that support previously learned capabilities, causing catastrophic
forgetting.
Meanwhile, prompts can accumulate case-specific rules
rather than generalizable principles, making new
requirements difficult to accommodate.
These limitations motivate EFRE, which checks historical
consistency before accepting updates and allows
functions to emerge when updates cannot be incorporated
into existing functions without compromising
historical performance.

\subsection{EFRE: Evolving Functional REpertoires}
\label{sec:efre}

We first define functions and their associated repertoire, then describe how the repertoire evolves during continual learning and supports inference.

\textbf{Functions and Functional Repertoire.}
A \emph{function} $f_k$ is a prompt encoding a learned
capability, such as a reasoning procedure or instructions
for handling particular inputs.
A \emph{functional repertoire}
$\mathcal{F}_t=\{f_1,\ldots,f_{K_t}\}$ contains the
$K_t$ functions available at stage $t$.
Each function maintains a history
$D_{\mathrm{history},k}$ of at most $H$ recent examples
successfully absorbed by that function.
Functions are not assigned to tasks in advance:
one function may serve multiple tasks, and one task
may involve multiple functions.
As learning proceeds, both individual functions and
the composition of the repertoire evolve.
For local evaluation, $R(f_k;D)$ denotes the performance
of the frozen LLM $\Phi$ on dataset $D$, directly
conditioned on $f_k$ and shared base instructions
$p_{\mathrm{base}}$.

\textbf{Repertoire Evolution and Inference.}
EFRE starts with one function and subsequently attempts
each update on the branch accepted by the most recent
successful update.
At stage $t$, the frozen LLM runs on
$D_{\mathrm{train},t}$ using the current function $f_k$.
The environment returns feedback $\mathcal{Q}_{t,k}$,
such as correctness scores, test results, or execution
errors, which the refinement model uses to propose
\begin{equation}
    f_k'
    =
    M_{\mathrm{refine}}
    (f_k,D_{\mathrm{train},t},\mathcal{Q}_{t,k}).
    \label{eq:function-refinement}
\end{equation}
Following Equation~\ref{eq:prompt-update}, the candidate must
improve performance on both $D_{\mathrm{train},t}$ and
$D_{\mathrm{val},t}$.
If either condition fails, it is discarded without
evaluating other functions.

A candidate that improves current performance must
also pass a \emph{historical consistency check}:
\begin{equation}
    R(f_k';D_{\mathrm{history},k})
    \geq
    (1-\epsilon)\,R(f_k;D_{\mathrm{history},k}),
    \label{eq:history-consistency}
\end{equation}
where scores are nonnegative and $\epsilon\in[0,1]$
specifies the maximum proportional decrease in
historical performance.
For example, $\epsilon=0.1$ requires retention of at
least $90\%$ of the original score.
The check is skipped when the history is empty.
A passing candidate replaces the current function,
and successfully absorbed examples are appended to
its history, retaining the most recent $H$ examples.

If the candidate improves current performance but
fails historical consistency, EFRE searches the
remaining functions in their stored order.
For each function, it obtains feedback on the incoming
batch and generates a revision using
Equation~\ref{eq:function-refinement}.
The search stops at the first candidate that improves
both training and validation performance relative to
its original function and passes
Equation~\ref{eq:history-consistency} on that function's
own history.
Only the matched function and its history change;
the original branch remains unchanged.
This enables \emph{functional reuse} through compatible
updates from different tasks.

If no compatible update is found, the refinement model
proposes a new function using the incoming data and
feedback from the failed attempts.
The new function is added only if it improves training
and validation performance relative to the current
prompt before this update.
Its history is initialized with up to $H$ examples
that it successfully handles.
We call this automatic creation of a new function
\emph{functional emergence}.
The repertoire can contain at most 50 functions;
when this limit is reached, an update that cannot be
absorbed by an existing function is discarded.
After an accepted refinement, reuse, or emergence,
the accepted branch becomes the current function
for the next update.
The repertoire thus evolves by incorporating compatible
capabilities into existing functions and creating new
ones when historical incompatibility prevents reuse.

At inference time, a textual selector $\sigma$ receives
the input and current function prompts, selects a
function, and directs the frozen LLM to generate a response:
\begin{equation}
    k(x)=\sigma(x;\mathcal{F}_t),
    \;
    \hat{y}
    =
    \Phi\!\left(x;p_{\mathrm{base}},f_{k(x)}\right).
    \label{eq:function-routing}
\end{equation}
The selector uses fixed instructions; its available
function prompts change as the repertoire evolves.
No task labels are required.
The shared instructions, repertoire, histories, and
selector constitute the external state $S_t$.
Learning updates this textual state while keeping
the parameters of $\Phi$ fixed.
Further methodological details are provided in Appendix~\ref{app:efre-details}.

\section{Experiments}
\label{sec:evaluation}

We evaluate EFRE on the ToolUse $\rightarrow$ FinQA
$\rightarrow$ SciKE-Bio sequence to assess its
continual-learning performance.
We conduct ablation studies and mechanism analyses
to examine historical consistency, function selection,
and functional specialization and reuse.
Finally, we evaluate EFRE on benchmarks such as HotpotQA to assess its effectiveness
in single-task optimization.

\begin{figure}[t]
    \centering
    \includegraphics[width=\linewidth]{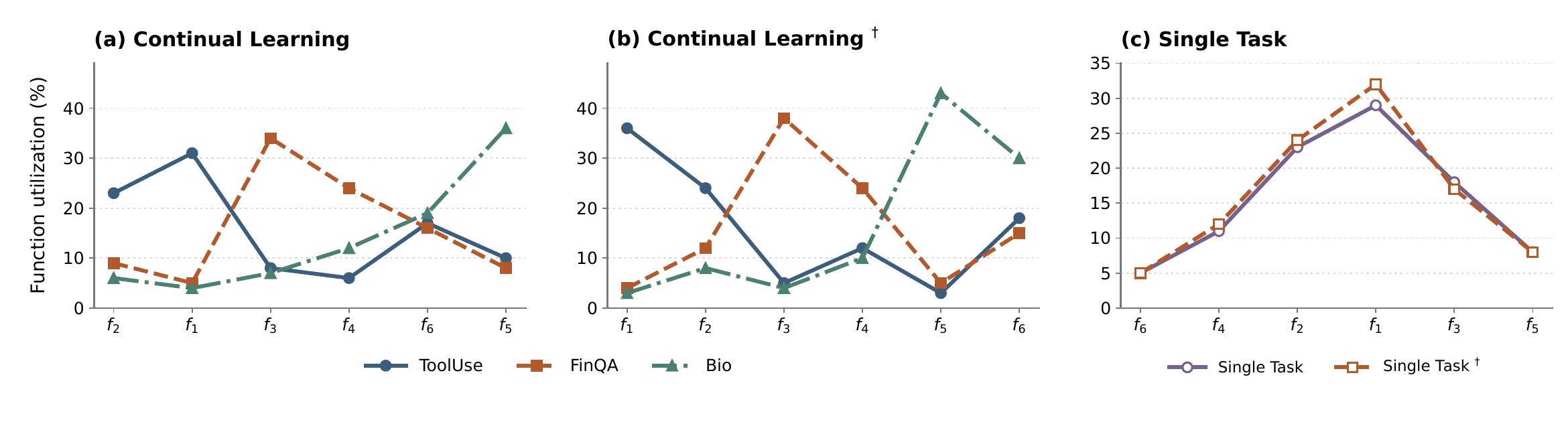}
    \vspace{-20pt}
    \caption{
    Functional reuse patterns in continual-learning and single-task settings.
    Curves show utilization of the six most-used functions.
    (a), (b) report utilization on ToolUse, FinQA,
    and SciKE-Bio.
    (c) compares both refinement configurations for
    single-task optimization, with a separate vertical scale.
    $\dagger$ denotes GPT-5.6-Luna as the refinement model;
    Qwen3-8B is used otherwise.
    }
    \label{fig:function-reuse}
\end{figure}

\begin{figure}[t]
    \centering
    \includegraphics[width=\linewidth]{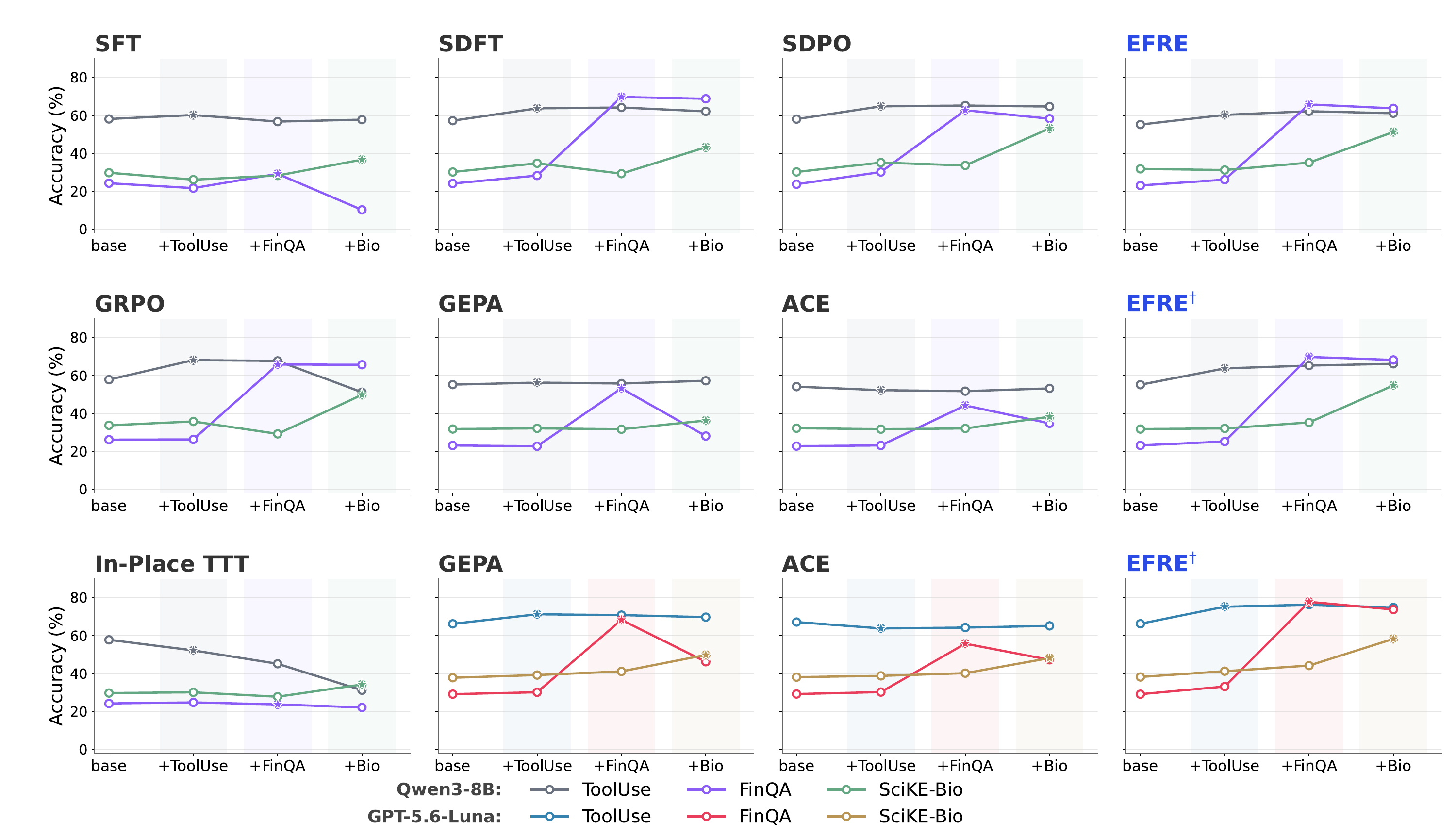}
    \caption{
    Continual-learning performance along the ToolUse $\rightarrow$ FinQA
    $\rightarrow$ SciKE-Bio sequence.
    Each panel shows one method, with curves tracking performance on
    all three tasks before adaptation and after each learning stage.
    Stars mark performance immediately after training on the corresponding
    task, and shaded regions indicate the task introduced at each stage.
    The two color palettes distinguish Qwen3-8B and GPT-5.6-Luna backbones.
    EFRE variants are grouped in the rightmost column;
    EFRE uses Qwen3-8B for refinement by defualt; $\dagger$ indicates GPT-5.6-Luna instead.
    }
    \label{fig:method-stages}
\end{figure}

\begin{table}[t]
\caption{
Functional organization of the learned repertoires.
For continual learning, \textbf{Cross-task reuse} denotes the
percentage of evaluation instances routed to functions
used by at least two tasks, averaged equally across ToolUse,
FinQA, and SciKE-Bio.
\textbf{Shared Func.} counts functions used by
at least two tasks, while \textbf{Func.} denotes total
repertoire size.
Cross-task reuse is not defined for single-task settings,
where each benchmark is optimized independently.
The final column illustrates a learned function.
Rows marked with $\dagger$ use GPT-5.6-Luna for refinement;
others use Qwen3-8B.
}
\label{tab:function_reuse}
\begin{center}
    \small
    \setlength{\tabcolsep}{2pt}
    \renewcommand{\arraystretch}{1.15}

    \begin{tabularx}{\linewidth}{@{}lcccX@{}}
        \toprule
        \textbf{Setting}
        & \shortstack{\textbf{Cross-task}\\\textbf{reuse}}
        & \shortstack{\textbf{Shared}\\\textbf{Func.}}
        & \textbf{Func.}
        & \textbf{Example learned function} \\
        \midrule

        Continual Learning
        & 41.7\%
        & 2
        & 7
        & Identify the next required action, select a tool,
          and check its arguments against the tool schema
          before execution. \\

        \rowcolor{efredefaultbg}
        Continual Learning$^{\dagger}$
        & \textbf{56.3\%}
        & \textbf{6}
        & 19
        & Extract the relevant financial values and units,
          write the calculation, and verify its result
          against the source table. \\

        \midrule

        Single-task
        & --
        & --
        & 8
        & Decompose the question into two linked subquestions, gather evidence, and synthesize a supported answer. \\

        \rowcolor{efredefaultbg}
        Single-task$^{\dagger}$
        & --
        & --
        & 12
        & Translate the conditions into equations, solve them, and verify candidates against the original constraints. \\

        \bottomrule
    \end{tabularx}
\end{center}
\end{table}

\subsection{Experimental Setup}
\label{sec:experimental_setup}

\textbf{Benchmark Datasets.}
We construct a continual-learning sequence using ToolUse~\citep{tang2023toolalpaca},
FinQA~\citep{chen2021finqa}, and SciKE-Bio~\citep{feng2024sciknoweval}.
After each learning stage, we evaluate performance on
the current and previously encountered tasks.
We additionally conduct single-task experiments on
HotpotQA~\citep{yang2018hotpotqa}, IFBench~\citep{pyatkin2025ifbench}, AIME-2025~\citep{maaAime}, and LiveBench-Math~\citep{white2024livebench}, and
evaluate continual learning in agent settings using
Continual Learning Bench~\citep{asawa2026clbench}.
We report the benchmark-specific evaluation metrics.
For continual-learning experiments, we also measure
plasticity and stability as defined in Section ~\ref{sec:problem},
which quantify improvement on the current task and
changes in performance on previous tasks, respectively.

\textbf{Comparing Baselines.}
We compare EFRE with methods that adapt model parameters
or external textual states.
Among parameter-based methods, SFT learns directly from
demonstrations through supervised fine-tuning.
SDFT~\citep{shenfeld2026selfdistillation} performs on-policy self-distillation using the model
conditioned on demonstrations as its teacher.
SDPO~\citep{hubotter2026sdpo} distills predictions from a feedback-conditioned
self-teacher into the policy.
GRPO~\citep{shao2024deepseekmath} performs reinforcement learning using relative
rewards within groups of sampled responses.
In-Place TTT~\citep{feng2026inplacettt} adapts selected MLP projection weights
during inference. Among methods that keep model parameters fixed,
GEPA~\citep{agrawal2025gepa} optimizes prompts through natural-language reflection
on execution traces and evaluation feedback, while
ACE~\citep{zhang2025ace} accumulates and refines contextual playbooks through
generation, reflection, and curation.
Detailed baseline configurations are provided in
Appendix~\ref{app:implementation}.

\textbf{Implementation Details.}
For the main experiments and agent continual-learning
experiments, we use Qwen3-8B~\citep{qwen2025eightbcard} and GPT-5.6-Luna~\citep{openai2026luna} as backbone
models and also use these two models for refinement.
For ablation studies and mechanism analyses, we use
Qwen3-8B as the backbone and GPT-5.6-Luna as the
refinement model.
For single-task experiments, we fix the backbone to
Qwen3-8B and evaluate Qwen3-8B and GPT-5.6-Luna
separately as refinement models.
All models are configured with low reasoning effort. Unless otherwise specified, we set the historical
consistency tolerance to $\epsilon=0.1$ and retain
the most recent $H=20$ successfully absorbed examples
for each function.
We use the refinement model with a dedicated routing prompt to select functions.
Single-task optimization runs once; other experiments are repeated three times, reporting mean and standard deviation.
Further implementation details are provided in Appendix~\ref{app:implementation}.

\begin{table}[t]
\caption{
Ablation studies on history retention, parameter settings,
and function selection along the ToolUse $\rightarrow$ FinQA
$\rightarrow$ Bio stream.
$\mathcal{J}_{\mathrm{plas}}$ measures the score change on the
current task, while $\mathcal{J}_{\mathrm{stab}}$ measures the
mean score change over previously encountered tasks with
equal weights.
Values are reported in percentage points; higher is better.
Stability is undefined at the first stage and denoted by --.
Baselines are shown in light beige, default configurations are shaded light blue, and the oracle
reference is shown in gray.
Bold means with pale gold shading indicate the best result
in each column within each ablation group, excluding
the baseline and oracle references.
}
\label{tab:ablation_studies}
\begin{center}
    \small
    \setlength{\tabcolsep}{4pt}
    \renewcommand{\arraystretch}{1.1}

    \begin{tabular}{@{}lcccccc@{}}
        \toprule
        \multirow[c]{2}{*}{\textbf{Setting}}
        & \multicolumn{2}{c}{\textbf{after ToolUse}}
        & \multicolumn{2}{c}{\textbf{after FinQA}}
        & \multicolumn{2}{c}{\textbf{after Bio}} \\
        \cmidrule(lr){2-3}
        \cmidrule(lr){4-5}
        \cmidrule(lr){6-7}

        & $\mathcal{J}_{\mathrm{plas}}\uparrow$
        & $\mathcal{J}_{\mathrm{stab}}\uparrow$
        & $\mathcal{J}_{\mathrm{plas}}\uparrow$
        & $\mathcal{J}_{\mathrm{stab}}\uparrow$
        & $\mathcal{J}_{\mathrm{plas}}\uparrow$
        & $\mathcal{J}_{\mathrm{stab}}\uparrow$ \\
        \midrule

        \rowcolor{baselinebg}
        GEPA
        & +1.07 \textpm{} 0.42
        & --
        & +30.48 \textpm{} 1.73
        & -0.49 \textpm{} 0.28
        & +4.59 \textpm{} 0.86
        & -11.82 \textpm{} 1.24 \\

        \rowcolor{baselinebg}
        GRPO
        & +10.28 \textpm{} 0.94
        & --
        & +39.46 \textpm{} 1.65
        & -0.40 \textpm{} 0.31
        & +20.58= \textpm{} 1.29
        & -8.32 \textpm{} 1.07 \\

        \midrule
        
        \multicolumn{7}{c}{
            \textbf{\textit{History Retention}}
        } \\
        \midrule

        \rowcolor{efredefaultbg}
        \textbf{Last-20}
        & \cellcolor{efrebestbg}\textbf{+8.60} \textpm{} 0.73
        & --
        & \cellcolor{efrebestbg}\textbf{+44.58} \textpm{} 1.36
        & +1.49 \textpm{} 0.24
        & \cellcolor{efrebestbg}\textbf{+19.52} \textpm{} 1.08
        & \cellcolor{efrebestbg}\textbf{-0.31} \textpm{} 0.19 \\

        Last-50
        & +6.76 \textpm{} 0.91 & --
        & +40.19 \textpm{} 1.62
        & +1.48 \textpm{} 0.31
        & +16.04 \textpm{} 1.27
        & -1.33 \textpm{} 0.38 \\

        Last-20+$D_{val}$
        & +7.69 \textpm{} 0.68 & --
        & +42.29 \textpm{} 1.45
        & \cellcolor{efrebestbg}\textbf{+1.59} \textpm{} 0.27
        & +17.85 \textpm{} 1.16
        & -0.81 \textpm{} 0.26 \\
        \midrule

        \multicolumn{7}{c}{
            \textbf{\textit{Tolerance $\epsilon$ Settings}}
        } \\
        \midrule

        0
        & +3.56 \textpm{} 0.57 & --
        & +31.57 \textpm{} 1.84
        & +0.88 \textpm{} 0.32
        & +11.46 \textpm{} 1.39
        & -3.78 \textpm{} 0.61 \\

        \rowcolor{efredefaultbg}
        \textbf{0.1}
        & \cellcolor{efrebestbg}\textbf{+8.60} \textpm{} 0.73
        & --
        & \cellcolor{efrebestbg}\textbf{+44.58} \textpm{} 1.36
        & \cellcolor{efrebestbg}\textbf{+1.49} \textpm{} 0.24
        & \cellcolor{efrebestbg}\textbf{+19.52} \textpm{} 1.08
        & \cellcolor{efrebestbg}\textbf{-0.31} \textpm{} 0.19 \\

        0.2
        & +7.46 \textpm{} 0.82 & --
        & +41.82 \textpm{} 1.51
        & +1.37 \textpm{} 0.29
        & +17.35 \textpm{} 1.22
        & -0.90 \textpm{} 0.34 \\

        0.5
        & +5.74 \textpm{} 0.76 & --
        & +37.52 \textpm{} 1.68
        & +1.26 \textpm{} 0.26
        & +15.29 \textpm{} 1.31
        & -1.44 \textpm{} 0.43 \\

        1.0
        & +4.53 \textpm{} 0.64 & --
        & +34.12 \textpm{} 1.79
        & +1.12 \textpm{} 0.35
        & +13.04 \textpm{} 1.46
        & -2.57 \textpm{} 0.52 \\
        \midrule

        \multicolumn{7}{c}{
            \textbf{\textit{Function Selection}}
        } \\
        \midrule

        First
        & +5.70 \textpm{} 0.88 & --
        & +38.40 \textpm{} 1.67
        & +1.07 \textpm{} 0.33
        & +15.43 \textpm{} 1.34
        & -1.61 \textpm{} 0.46 \\

        Random
        & +5.25 \textpm{} 1.12 & --
        & +37.61 \textpm{} 2.04
        & +1.18 \textpm{} 0.41
        & +15.17 \textpm{} 1.63
        & -1.44 \textpm{} 0.57 \\

        Embedding-20
        & +5.84 \textpm{} 0.79 & --
        & +38.59 \textpm{} 1.58
        & +1.35 \textpm{} 0.28
        & +15.75 \textpm{} 1.25
        & -1.45 \textpm{} 0.39 \\

        Embedding-50
        & +8.34 \textpm{} 0.77 & --
        & +44.30 \textpm{} 1.42
        & +1.46 \textpm{} 0.26
        & +19.31 \textpm{} 1.14
        & \cellcolor{efrebestbg}\textbf{-0.29} \textpm{} 0.22 \\

        \rowcolor{efredefaultbg}
        \textbf{LLM-selector}
        & \cellcolor{efrebestbg}\textbf{+8.60} \textpm{} 0.73
        & --
        & \cellcolor{efrebestbg}\textbf{+44.58} \textpm{} 1.36
        & \cellcolor{efrebestbg}\textbf{+1.49} \textpm{} 0.24
        & \cellcolor{efrebestbg}\textbf{+19.52} \textpm{} 1.08
        & -0.31 \textpm{} 0.19 \\

        \textcolor{efreoraclegray}{\textbf{Oracle}}
        & \textcolor{efreoraclegray}{+12.06 \textpm{} 0.65}
        & \textcolor{efreoraclegray}{--}
        & \textcolor{efreoraclegray}{+50.76 \textpm{} 1.21}
        & \textcolor{efreoraclegray}{+1.94 \textpm{} 0.23}
        & \textcolor{efreoraclegray}{+24.35 \textpm{} 0.97}
        & \textcolor{efreoraclegray}{-0.32 \textpm{} 0.18} \\

        \bottomrule
    \end{tabular}
\end{center}
\end{table}

\subsection{Main Results}

\textbf{Observation 1: EFRE achieves strong continual-learning
performance and exhibits functional reuse and specialization.}
We evaluate Qwen3-8B and GPT-5.6-Luna as backbone models;
$\dagger$ denotes GPT-5.6-Luna refinement, while variants
without $\dagger$ use Qwen3-8B.
Figure~\ref{fig:method-stages} shows that EFRE learns new
tasks while largely preserving earlier gains.
With Qwen3-8B, EFRE$^{\dagger}$ loses only 1.56 percentage
points on FinQA after Bio adaptation, compared with
25.10 for GEPA and 9.43 for ACE, and reaches final
ToolUse/FinQA/Bio scores of $66.21/68.26/54.83$.
With GPT-5.6-Luna, it retains $73.76\%$ FinQA accuracy,
versus $46.24\%$ for GEPA and $47.28\%$ for ACE.
Together, these results show a stronger balance between
new-task adaptation and retention of previously learned
capabilities. Figures~\ref{fig:function-reuse}(a,b) further show that
the learned repertoire combines specialization with
cross-task reuse.
For example, in panel~(a), $f_1$, $f_3$, and $f_5$
predominantly serve ToolUse, FinQA, and Bio, respectively,
whereas $f_6$ is shared across all three tasks, with
utilization rates of $17\%$, $16\%$, and $19\%$.
Table~\ref{tab:function_reuse} quantifies this behavior
through cross-task reuse and the number of shared functions,
while the example prompts illustrate distinct functional
roles.
Thus, EFRE does not simply allocate an independent prompt to each task, but develops specialized and reusable functional contexts.

\subsection{Ablation Study and Mechanism Analysis}

\textbf{Observation 2: Historical consistency improves
continual-learning performance by better balancing
plasticity and stability.}
Table~\ref{tab:ablation_studies} varies history retention
and tolerance separately, holding other hyperparameters fixed.
We compare the most recent 20 examples (Last-20),
50 examples (Last-50), and 20 examples supplemented with
validation data (Last-20+$D_{\mathrm{val}}$), using
plasticity and stability defined in
Equation.~\ref{eq:plasticity} and~\ref{eq:stability}.
Last-20 yields the highest plasticity at every stage
and the least historical degradation after Bio,
showing that more history does not necessarily improve learning.
Among the tested tolerances, $\epsilon=0.1$ achieves
the best plasticity and stability at every applicable stage.
Its Bio-stage plasticity is +19.52 points, versus +11.46
with strict preservation ($\epsilon=0$) and +13.04
without a historical performance constraint ($\epsilon=1$).
It exceeds GEPA (+4.59) and approaches GRPO (+20.58),
with stability of -0.31 points, versus
-11.82 and -8.32, respectively.
These results favor a moderate constraint supporting
new learning while preserving earlier capabilities.

\begin{table}[t]
\caption{
Performance comparison across four benchmarks.
Both EFRE configurations use Qwen3-8B as the backbone.
EFRE and EFRE$^{\dagger}$ use Qwen3-8B and GPT-5.6-Luna
as the refinement model, respectively.
Aggregate is the unweighted mean across benchmarks,
and Improvement is the percentage-point gain over Base.
EFRE and EFRE$^{\dagger}$ are shaded light blue and darker blue,
respectively.
Bold means indicate the best result in each column.
}
\label{tab:main-results}
\begin{center}
    \small
    \setlength{\tabcolsep}{3pt}
    \renewcommand{\arraystretch}{1.1}
    \begin{tabular}{@{}lcccccc@{}}
        \toprule
        \textbf{Method}
        & \textbf{HotpotQA}
        & \textbf{IFBench}
        & \textbf{AIME-2025}
        & \shortstack{\textbf{LiveBench}\\\textbf{Math}}
        & \textbf{Aggregate}
        & \textbf{Improvement} \\
        \midrule

        Base
        & 42.33 & 36.90 & 27.33 & 48.70
        & 38.82 & -- \\
        \midrule

        GRPO
        & 43.33 & 35.88 & \textbf{38.00} & 51.26
        & 42.12 & +3.30 \\

        ACE
        & 60.00 & 35.72 & 31.00 & 52.76
        & 44.87 & +6.06 \\

        GEPA
        & 62.33 & 38.61 & 32.00 & 51.95
        & 46.22 & +7.41 \\

        GEPA+Merge
        & 64.33 & 28.23 & 32.00 & 51.95
        & 44.13 & +5.31 \\
        \midrule

        \rowcolor{efredefaultbg}
        \textbf{EFRE}
        & 70.67
        & 39.26
        & 34.33
        & 51.88
        & 49.04
        & +10.22 \\

        \rowcolor{efreteacherbg}
        \textbf{EFRE$^{\dagger}$}
        & \textbf{72.33}
        & \textbf{42.38}
        & 36.00
        & \textbf{54.32}
        & \textbf{51.26}
        & \textbf{+12.44}\\

        \bottomrule
    \end{tabular}
\end{center}
\end{table}

\textbf{Observation 3: LLM-based selection enables
effective use of the functional repertoire.}
Table~\ref{tab:ablation_studies} compares selection
strategies with other hyperparameters fixed.
First always selects the first function, while Random
samples uniformly.
Embedding-20 and Embedding-50 use Qwen3-Embedding-0.6B
fine-tuned on 20 and 50 historical examples per function
for retrieval.
The LLM selector chooses from the function prompts
based on the input, whereas Oracle evaluates all
functions and selects the highest-scoring output,
providing an upper-bound reference within the repertoire.
The LLM selector achieves the highest non-oracle
plasticity at every stage and better stability than
First and Random.
Embedding-50 yields slightly better Bio-stage stability
(-0.29 versus -0.31 points), but lower plasticity throughout.
Oracle further improves Bio-stage plasticity from
+19.52 to +24.35 with comparable stability,
indicating room to improve selection within the
existing repertoire.

\textbf{Observation 4: Functional repertoires also
benefit single-task optimization.}
Table~\ref{tab:main-results} shows that EFRE achieves
an aggregate score of 49.04 across four benchmarks,
compared with 46.22 for GEPA, 44.87 for ACE, and
42.12 for GRPO.
Using GPT-5.6-Luna for refinement further improves
the aggregate score to 51.26.
Figure~\ref{fig:function-reuse}(c) shows that, even
within independently optimized single-task settings,
utilization is distributed across multiple functions
rather than concentrated in a single context.
Table~\ref{tab:function_reuse} reports repertoires of
8 and 12 functions under the two refinement settings,
with example functions capturing distinct procedures
such as evidence decomposition and equation-based
reasoning.
These results suggest that functional differentiation
is useful beyond continual learning: multiple specialized
procedures can emerge within a single task even without
task transitions.

\begin{table}[t]
\caption{
Minimal agent results across six interactive environments
with two backbones.
Each backbone uses ICL, GEPA, ACE, or EFRE.
Values are mean \textpm{} standard deviation; higher is better.
EFRE rows are light blue.
Bold means indicate the best result within each backbone,
including ties.
}
\label{tab:agent}
\begin{center}
    \small
    \setlength{\tabcolsep}{3pt}
    \renewcommand{\arraystretch}{1.1}

    \begin{tabular}{@{}lcccccc@{}}
        \toprule
        \textbf{Method}
        & \shortstack{\textbf{Blind Spectrum}\\\textbf{Monitoring}}
        & \shortstack{\textbf{Codebase}\\\textbf{Adaptation}}
        & \shortstack{\textbf{Cohort}\\\textbf{Studies}}
        & \shortstack{\textbf{Database}\\\textbf{Exploration}}
        & \shortstack{\textbf{Exploitable}\\\textbf{Poker}}
        & \shortstack{\textbf{Sales}\\\textbf{Prediction}} \\
        \midrule

        \multicolumn{7}{c}{\textbf{\textit{Qwen3-8B}}} \\
        \midrule

        + ICL
        & 12.0 \textpm{} 5.6
        & -32.0 \textpm{} 8.4
        & -28.0 \textpm{} 7.9
        & 10.0 \textpm{} 5.1
        & -8.0 \textpm{} 6.7
        & 25.0 \textpm{} 6.2 \\

        + GEPA
        & 26.0 \textpm{} 4.8
        & -15.0 \textpm{} 7.2
        & -18.0 \textpm{} 6.8
        & 22.0 \textpm{} 4.6
        & 0.0 \textpm{} 5.9
        & 42.0 \textpm{} 5.3 \\

        + ACE
        & 2.0 \textpm{} 4.4
        & -10.0 \textpm{} 6.9
        & -24.0 \textpm{} 7.3
        & 5.0 \textpm{} 4.8
        & -5.0 \textpm{} 5.6
        & 15.0 \textpm{} 5.9 \\

        \rowcolor{efredefaultbg}
        \textbf{+ EFRE}
        & \textbf{35.0} \textpm{} 4.1
        & \textbf{-5.0} \textpm{} 6.2
        & \textbf{-10.0} \textpm{} 6.0
        & \textbf{30.0} \textpm{} 4.2
        & \textbf{6.0} \textpm{} 5.1
        & \textbf{52.0} \textpm{} 4.7 \\
        \midrule

        \multicolumn{7}{c}{\textbf{\textit{GPT-5.6-Luna}}} \\
        \midrule

        + ICL
        & 36.0 \textpm{} 4.3
        & 12.0 \textpm{} 6.4
        & 0.0 \textpm{} 5.8
        & 27.0 \textpm{} 4.5
        & -2.0 \textpm{} 5.3
        & 60.0 \textpm{} 4.8 \\

        + GEPA
        & 44.0 \textpm{} 3.8
        & 18.0 \textpm{} 5.7
        & 4.0 \textpm{} 5.4
        & 35.0 \textpm{} 4.0
        & 8.0 \textpm{} 4.8
        & 68.0 \textpm{} 4.2 \\

        + ACE
        & 8.0 \textpm{} 4.1
        & \textbf{25.0} \textpm{} 5.9
        & -7.0 \textpm{} 6.1
        & 12.0 \textpm{} 4.4
        & 2.0 \textpm{} 5.0
        & 18.0 \textpm{} 5.2 \\

        \rowcolor{efredefaultbg}
        \textbf{+ EFRE}
        & \textbf{52.0} \textpm{} 3.5
        & \textbf{25.0} \textpm{} 5.2
        & \textbf{8.0} \textpm{} 4.9
        & \textbf{42.0} \textpm{} 3.7
        & \textbf{15.0} \textpm{} 4.4
        & \textbf{76.0} \textpm{} 3.9 \\

        \bottomrule
    \end{tabular}
\end{center}
\end{table}

\section{Extension: EFRE for Agent Continual Learning}
\label{sec:agent}

Motivated by EFRE's effectiveness in LLM continual learning,
we extend it to a minimal agent system to examine whether
its learning mechanism remains effective when training data
arise from environmental interactions.

\textbf{A Minimal Continual-Learning Agent.}
We adopt the lightweight scaffold from Continual Learning
Bench~\citep{asawa2026clbench} and integrate all methods into the same
agent--environment interaction loop.
In EFRE, a single frozen LLM performs task execution,
function selection, and refinement.
Given the current observation, it selects a function
and generates actions under its prompt.
Interaction feedback guides functional updates,
while historical consistency checks govern their
acceptance, reuse in another branch, or functional emergence.
Adaptation updates the external repertoire while keeping
the selector instructions and model parameters fixed.

\textbf{Evaluation Results.}
We evaluate six tasks from Continual Learning Bench:
Blind Spectrum Monitoring, Codebase Adaptation,
Cohort Studies, Database Exploration, Exploitable Poker,
and Sales Prediction.
Table~\ref{tab:agent} shows that EFRE achieves the highest
mean score on all six tasks with Qwen3-8B.
Its unweighted average is 18.0, versus 9.5 for
GEPA~\citep{agrawal2025gepa} and -3.5 for
ICL~\citep{brown2020fewshot}.
Gains include Codebase Adaptation (-5.0 versus GEPA's
-15.0) and Sales Prediction (52.0 versus 42.0).
With GPT-5.6-Luna~\citep{openai2026luna}, EFRE averages
36.3, versus 29.5 for GEPA and 22.2 for ICL.
It exceeds the strongest baseline on five tasks and
matches ACE on Codebase Adaptation.
These gains across environments and backbones extend
EFRE's benefits to agent continual learning through
learning from ongoing environmental interactions.

\section{Conclusions}
We introduced EFRE, a non-parametric continual learning
method that evolves a repertoire of functions represented
by prompts while keeping the underlying LLM fixed.
Historical consistency checks guide the refinement and
reuse of existing functions and the emergence of new ones,
allowing the repertoire to adapt as learning tasks arrive.
Experiments show that EFRE reduces forgetting and improves
overall continual-learning performance, with functional
specialization and reuse also benefiting single-task
optimization. Ablation studies support the roles of historical
consistency and function selection, while an extension
to a minimal agent system demonstrates gains in
interactive environments.
Together, these findings suggest that evolving functional
repertoires offer a promising approach to continual
learning without model parameter updates. Moreover, EFRE can be naturally extended to agent settings, where its evolving functional repertoire supports continual adaptation from ongoing interactions with dynamic environments.

\bibliography{references}
\bibliographystyle{iclr2027_conference}

\clearpage

\appendix

\section{Datasets, Benchmarks, and Models}
\label{app:datasets-models}

\subsection{Datasets and Benchmarks}

\textbf{LLM Continual Learning.}
We use the ToolUse, FinQA, and SciKE-Bio datasets and
evaluation protocols provided by
\citep{harrington2026continual}, following the sequence
ToolUse $\rightarrow$ FinQA $\rightarrow$ SciKE-Bio.

\begin{itemize}
    \item \textbf{ToolUse.}
    ToolUse evaluates tool selection and argument construction
    from natural-language requests.
    It is derived from ToolAlpaca, which contains 3,938 simulated
    tool-use instances covering over 400 APIs across
    50 categories~\citep{tang2023toolalpaca}.
    Each input provides a request and relevant tool information,
    and the model produces the corresponding tool calls.
    We use it to study the acquisition and retention of
    tool-use capabilities.

    \item \textbf{FinQA.}
    FinQA is a financial question-answering dataset containing
    expert-annotated questions over financial reports, together
    with numerical answers and reasoning
    programs~\citep{chen2021finqa}.
    Questions require integrating evidence from text and tables
    and performing calculations involving quantities such as
    percentages, changes, and ratios.
    We use it to study continual learning of financial
    numerical reasoning.

    \item \textbf{SciKE-Bio.}
    SciKE-Bio is a multiple-choice biology subset derived
    from SciKnowEval, a benchmark covering scientific knowledge
    in biology, chemistry, physics, and materials
    science~\citep{feng2024sciknoweval}.
    As the final task in the sequence, it allows us to examine
    scientific knowledge acquisition and its effects on
    previously learned capabilities.
\end{itemize}

\textbf{Single-Task Optimization.}
We use four benchmarks to assess whether EFRE also improves
learning within individual tasks.

\begin{itemize}
    \item \textbf{HotpotQA.}
    HotpotQA is a Wikipedia-based multi-hop question-answering
    benchmark with annotated answers and supporting
    facts~\citep{yang2018hotpotqa}.
    Its questions require combining evidence across documents,
    including identifying connections between entities and
    comparing their attributes.
    We use it to assess multi-hop reasoning and
    evidence-based question answering.

    \item \textbf{IFBench.}
    IFBench evaluates instruction following through
    automatically verifiable constraints, including output
    formatting, lexical requirements, and combinations of
    instructions~\citep{pyatkin2025ifbench}.
    It emphasizes generalization to constraints beyond those
    encountered during training.
    We use it to assess the model's ability to follow
    diverse instructions.

    \item \textbf{AIME-2025.}
    AIME-2025 contains the 30 problems from the two 2025
    American Invitational Mathematics
    Examinations~\citep{maaAime}.
    These problems require competition-level mathematical
    reasoning and have integer answers.
    We use it to assess mathematical problem solving.

    \item \textbf{LiveBench-Math.}
    LiveBench-Math is the mathematical subset of LiveBench,
    a regularly updated benchmark with automatically
    checkable answers~\citep{white2024livebench}.
    We use it to assess mathematical reasoning on an
    additional set of problems.
\end{itemize}

\textbf{Agent Continual Learning.}

\begin{itemize}
    \item \textbf{Continual Learning Bench.}
    Continual Learning Bench evaluates continual adaptation
    through sequential interactions in stateful
    environments~\citep{asawa2026clbench}.
    Its tasks require agents to accumulate information,
    learn recurring environmental structure, and adapt
    to changes across interactions.
    We evaluate six environments: Blind Spectrum Monitoring,
    Codebase Adaptation, Cohort Studies, Database Exploration,
    Exploitable Poker, and Sales Prediction.
    Together, they test whether EFRE supports continual
    learning from agent--environment interactions.
\end{itemize}

\subsection{Models}
\label{app:models}

\begin{itemize}
    \item \textbf{Qwen3-Embedding-0.6B.}
    Qwen3-Embedding-0.6B is an instruction-aware text embedding
    model with approximately 0.6 billion parameters, a 32K-token
    context window, and configurable embedding dimensions
    from 32 to 1,024~\citep{qwen2025embeddingcard}.
    We use it for the Embedding-20 and Embedding-50
    function-selection baselines, fine-tuning on 20 and
    50 historical examples per function, respectively.

    \item \textbf{Qwen3-8B.}
    Qwen3-8B is an open-weight language model with 8.2 billion
    parameters and a native context window of 32,768 tokens,
    supporting both thinking and non-thinking
    modes~\citep{qwen2025eightbcard}.
    We use it as a backbone, refinement model, and function
    selector in the corresponding experimental configurations.
    Its parameters remain fixed under EFRE.

    \item \textbf{GPT-5.6-Luna.}
    GPT-5.6-Luna is an API-served language model supporting
    reasoning, tool calling, and structured
    outputs~\citep{openai2026luna}.
    We use it as a backbone, refinement model, and function
    selector, with reasoning effort set to Low.
    EFRE$^{\dagger}$ uses GPT-5.6-Luna for refinement,
    whereas EFRE without $\dagger$ uses Qwen3-8B.
\end{itemize}

\section{Implementation Details}
\label{app:implementation}

\subsection{Hyperparameters}
\label{app:hyperparameters}

Table~\ref{tab:app-efre-hyperparameters} summarizes the default
EFRE configuration. Unless otherwise specified, we use a
micro-batch size of 3, retain the 20 most recent successfully
absorbed examples per function, and set the historical
consistency tolerance to 0.1. The maximum repertoire size is 50.
The backbone and refinement models follow the configurations
reported in each experiment. All experiments are repeated
three times.

\begin{table}[t]
\caption{Default EFRE hyperparameters. Model choices correspond
to the experimental configurations reported in the main text.}
\label{tab:app-efre-hyperparameters}
\begin{center}
\small
\setlength{\tabcolsep}{6pt}
\renewcommand{\arraystretch}{1.1}
\begin{tabular}{@{}ll@{}}
\toprule
\textbf{Hyperparameter} & \textbf{Setting} \\
\midrule
Micro-batch size & 3 \\
History capacity $H$ & 20 examples per function \\
Historical consistency tolerance $\epsilon$ & 0.1 \\
Maximum number of functions & 50 \\
Backbone model & Qwen3-8B, GPT-5.6-Luna \\
Refinement model & Qwen3-8B, GPT-5.6-Luna \\
Backbone reasoning effort & Low \\
Refinement reasoning effort & Low \\
Embedding model & Qwen3-Embedding-0.6B \\
\bottomrule
\end{tabular}
\end{center}
\end{table}

Table~\ref{tab:app-training-config} summarizes optimizer and
batch-size settings for the comparison methods. Parameter-updating
methods use AdamW~\citep{loshchilov2019adamw} with a batch size of 16; GEPA and ACE use
batches of 3 examples for textual updates. The Embedding-20 and
Embedding-50 ablations restrict embedding-model training to
20 and 50 historical examples per function, respectively.

\begin{table}[t]
\caption{Optimizer and batch-size settings for comparison methods.
A dash indicates that the setting is not applicable.}
\label{tab:app-training-config}
\begin{center}
\small
\setlength{\tabcolsep}{3pt}
\renewcommand{\arraystretch}{1.1}
\begin{tabular}{@{}lcc@{}}
\toprule
\textbf{Method} & \textbf{Optimizer} & \textbf{Batch size} \\
\midrule
SFT & AdamW & 16 \\
SDFT & AdamW & 16 \\
SDPO & AdamW & 16 \\
GRPO & AdamW & 16 \\
In-Place TTT & AdamW & 16 \\
GEPA & -- & 3 \\
ACE & -- & 3 \\
ICL & -- & -- \\
\bottomrule
\end{tabular}
\end{center}
\end{table}

\subsection{Dataset Setup}
\label{app:dataset-setup}

For LLM continual learning, we follow the dataset setup of
\citep{harrington2026continual}, with tasks arriving in the order
ToolUse $\rightarrow$ FinQA $\rightarrow$ SciKE-Bio.
Each single-task benchmark is optimized independently.
Training, validation, and evaluation splits are disjoint and
shared across methods. Training data provide optimization
examples and feedback, validation data support update selection,
and evaluation data measure performance. Function histories
are drawn from training data; the Last-20+$D_{\mathrm{val}}$
ablation additionally uses validation data for consistency checks.
For AIME-2025, training and validation questions come from
2022--2024, while evaluation uses the 30 questions from 2025. Agent continual learning follows ordered interactions in
Continual Learning Bench rather than a static three-way split.
Updates use only information available from completed
interactions.

\subsection{Evaluation Protocols}
\label{app:evaluation-protocols}

\textbf{Task Scoring.}
We evaluate each benchmark on its held-out evaluation set
$D_{\mathrm{eval}}$ using the official scoring criteria.
Evaluation examples are excluded from training, prompt
refinement, and update selection.
The same evaluation sets and scoring procedures are used
across methods.

\textbf{Evaluation Matrix.}
Let $T$ denote the number of tasks in the continual-learning
sequence, $S_0$ the initial learner state, and $S_t$ the state
after learning task $t$.
For EFRE, $S_t$ includes the functional repertoire and
selector, while the backbone parameters remain fixed.
For parameter-updating baselines, evaluation uses the model
checkpoint obtained after the corresponding stage.

We evaluate every task before learning begins and after
each learning stage, forming the performance matrix
\begin{equation}
    A_{t,i}
    =
    R(S_t;D_{\mathrm{eval},i}),
    \;
    0 \leq t \leq T,\quad 1 \leq i \leq T,
    \label{eq:app-score-matrix}
\end{equation}
where $A_{t,i}$ is the score on task $i$ after stage $t$.
Each row therefore describes one learner state across all
tasks, and each column tracks performance on one task
throughout learning.
Row $t=0$ contains the initial scores before any adaptation.

Evaluation leaves the learner unchanged.
For EFRE, each input is routed by the function selector,
and the selected function is used to generate the response.
Reported scores thus evaluate the complete system,
including function selection.
Oracle selection is used only in the explicitly designated
oracle ablation.
For the LLM continual-learning experiments, we express
$A_{t,i}$ on a 0--100 scale; differences between these scores
are reported in percentage points.

\textbf{Plasticity.}
Plasticity measures how much learning at stage $t$ improves
performance on the current task:
\begin{equation}
    \mathcal{J}_{\mathrm{plas}}^{(t)}
    =
    A_{t,t}-A_{t-1,t}.
    \label{eq:app-plasticity}
\end{equation}
Here, $A_{t-1,t}$ is the score on task $t$ immediately before
learning it, and $A_{t,t}$ is the score immediately afterward.
The comparison therefore isolates the performance change
during the current stage, including when earlier learning
has already affected the current task.
A positive value indicates improvement, zero indicates
no change, and a negative value indicates degradation.

For the ToolUse $\rightarrow$ FinQA $\rightarrow$ SciKE-Bio
sequence, the three plasticity values are
\begin{equation}
    \begin{aligned}
        \mathcal{J}_{\mathrm{plas}}^{(1)}
        &= A_{1,1}-A_{0,1},\\
        \mathcal{J}_{\mathrm{plas}}^{(2)}
        &= A_{2,2}-A_{1,2},\\
        \mathcal{J}_{\mathrm{plas}}^{(3)}
        &= A_{3,3}-A_{2,3}.
    \end{aligned}
    \label{eq:app-three-plas}
\end{equation}
For example, FinQA plasticity compares its score after
FinQA adaptation with its score after ToolUse adaptation,
rather than with the initial model's FinQA score.

\textbf{Stability.}
Stability measures how learning at stage $t$ changes
performance on previously learned tasks:
\begin{equation}
    \mathcal{J}_{\mathrm{stab}}^{(t)}
    =
    \frac{1}{t-1}
    \sum_{i=1}^{t-1}
    \left(A_{t,i}-A_{t-1,i}\right),
    \; t>1.
    \label{eq:app-stability}
\end{equation}
For each previous task, we compare its scores immediately
before and after the current stage, then average these
changes with equal weight across tasks.
This weighting prevents tasks with larger evaluation sets
from dominating the measure.

A negative value indicates average forgetting, zero
indicates no net change in previous-task performance,
and a positive value indicates average improvement.
Because the measure averages across tasks, gains on one
task can offset losses on another.
We therefore report individual task scores alongside
the aggregate stability measure.
Stability is undefined at the first stage, when no
previous tasks exist, and is denoted by ``--''.

For the three-task sequence,
\begin{equation}
    \begin{aligned}
        \mathcal{J}_{\mathrm{stab}}^{(2)}
        &= A_{2,1}-A_{1,1},\\
        \mathcal{J}_{\mathrm{stab}}^{(3)}
        &= \frac{1}{2}
        \left[
            (A_{3,1}-A_{2,1})
            +(A_{3,2}-A_{2,2})
        \right].
    \end{aligned}
    \label{eq:app-three-stab}
\end{equation}
Thus, stability after FinQA measures the change on
ToolUse, while stability after SciKE-Bio averages the
changes on ToolUse and FinQA.
This definition measures the effect of the current
learning stage. It differs from cumulative backward
transfer~\citep{lopezpaz2017gem}, which compares each previous task's current
score with its score immediately after that task
was originally learned.

\textbf{Aggregation and Uncertainty.}
Final continual-learning performance is the unweighted
mean of all task scores after the last stage:
\begin{equation}
    \overline{A}_{\mathrm{final}}
    =
    \frac{1}{T}\sum_{i=1}^{T}A_{T,i}.
    \label{eq:app-final-average}
\end{equation}
For single-task optimization, we similarly report the
unweighted mean of the four benchmark scores.
Each benchmark contributes equally to the aggregate,
regardless of its evaluation-set size.

All experiments are repeated three times.
We first compute task aggregates, plasticity, and
stability separately within each run, and then calculate
their means and sample standard deviations across runs.
For any reported quantity $z$, let $z^{(r)}$ denote its
value in run $r$. We compute
\begin{equation}
    \bar{z}
    =
    \frac{1}{3}\sum_{r=1}^{3}z^{(r)},
    \;
    s_z
    =
    \sqrt{
        \frac{1}{2}
        \sum_{r=1}^{3}
        \left(z^{(r)}-\bar{z}\right)^2
    }.
    \label{eq:app-run-sd}
\end{equation}
Results with uncertainty estimates are reported as
$\bar{z}\pm s_z$.
For plasticity and stability, the before-and-after
scores are paired within the same run before computing
the standard deviation.
Their uncertainty cannot be recovered from the separate
checkpoint standard deviations alone, because the scores
within a run may be correlated.
Tables calculated from aggregate checkpoint scores
therefore report means only.
Small rounding differences may arise when calculations
use displayed scores rather than the original
unrounded values.

\textbf{Agent Evaluation.}
We evaluate agents through the ordered interaction
schedule of each environment in Continual Learning
Bench~\citep{asawa2026clbench}.
For environment $e$, let $r_{e,j}$ be the native reward
on interaction instance $j$, and let $N_e$ be the number
of evaluated instances.
The environment-level mean reward is
\begin{equation}
    \bar{r}_e
    =
    \frac{1}{N_e}\sum_{j=1}^{N_e}r_{e,j}.
    \label{eq:app-agent-mean-reward}
\end{equation}
Native reward definitions differ across environments.
We therefore report the benchmark's normalized reward:
\begin{equation}
    \widehat{r}_e
    =
    100\,
    \frac{
        \bar{r}_e-r_e^{\mathrm{ref}}
    }{
        r_e^{\max}-r_e^{\mathrm{ref}}
    },
    \label{eq:app-agent-normalization}
\end{equation}
where $r_e^{\mathrm{ref}}$ is the benchmark's fixed
reference reward and $r_e^{\max}$ is its designated
upper reference value, with
$r_e^{\max}>r_e^{\mathrm{ref}}$.
The same normalization constants are used for all
methods evaluated in that environment.
A normalized reward of zero matches the reference,
100 matches the upper reference value, and a negative
value indicates performance below the reference.
These scores are normalized rewards, not accuracy
percentages.

Table~\ref{tab:agent} reports normalized reward for each
environment. The overall agent score assigns equal
weight to the six environments:
\begin{equation}
    \overline{r}_{\mathrm{agent}}
    =
    \frac{1}{6}\sum_{e=1}^{6}\widehat{r}_e.
    \label{eq:app-agent-average}
\end{equation}
Normalization and aggregation are performed within
each run before summarizing results across the
three runs.

Normalized reward differs from normalized gain, which
compares an adapting agent with its own stateless
counterpart:
\begin{equation}
    \widehat{g}_e
    =
    100\,
    \frac{
        \bar{r}_e^{\mathrm{stateful}}
        -
        \bar{r}_e^{\mathrm{stateless}}
    }{
        r_e^{\max}
        -
        \bar{r}_e^{\mathrm{stateless}}
    }.
    \label{eq:app-agent-normalized-gain}
\end{equation}
Normalized gain requires a separate stateless evaluation
and a positive denominator.
Our agent results report normalized reward;
normalized gain is not used.
Neither quantity is the stage-local stability measure
$\mathcal{J}_{\mathrm{stab}}$ defined for the sequential
LLM tasks above.
\section{Additional Results}
\label{app:additional-results}

\subsection{Full Results}
\label{app:full-results}

Tables~\ref{tab:app-full-scores} and~\ref{tab:app-full-deltas}
report stage-wise performance for all 12 configurations.
EFRE$^{\dagger}$ uses GPT-5.6-Luna for refinement, whereas
unmarked EFRE uses Qwen3-8B. With Qwen3-8B as the backbone,
EFRE$^{\dagger}$ achieves a final average of 63.10, exceeding
GRPO and GEPA by 7.50 and 22.52 points, respectively.
Its Bio-stage plasticity is +19.52, close to GRPO's +20.58,
while its stability is substantially higher (-0.31 versus
-8.32). Using Qwen3-8B for refinement also yields a final
average of 58.75 with Bio-stage stability of -1.57.
With GPT-5.6-Luna as the backbone, EFRE$^{\dagger}$ reaches
68.96, compared with 55.28 for GEPA and 53.54 for ACE.
These results show that EFRE combines effective adaptation
with stronger retention across both backbones.

\begin{table}[t]
\caption{Full continual-learning results. Each stage reports ToolUse, FinQA,
and SciKE-Bio accuracy (\%); Bio denotes SciKE-Bio. Final Avg. is the
unweighted mean after Bio. Bold values indicate the highest score in
each column within each backbone, including ties. Light and dark blue rows denote EFRE and EFRE$^{\dagger}$,
respectively.}
\label{tab:app-full-scores}
\begin{center}
\small
\setlength{\tabcolsep}{3pt}
\renewcommand{\arraystretch}{1.1}
\resizebox{\linewidth}{!}{%
\begin{tabular}{@{}lccccccccccccc@{}}
\toprule
\textbf{Method} & \multicolumn{3}{c}{\textbf{Base}} & \multicolumn{3}{c}{\textbf{After ToolUse}} & \multicolumn{3}{c}{\textbf{After FinQA}} & \multicolumn{3}{c}{\textbf{After Bio}} & \textbf{Final Avg.} \\
\cmidrule(lr){2-4}\cmidrule(lr){5-7}\cmidrule(lr){8-10}\cmidrule(lr){11-13}
 & \textbf{ToolUse} & \textbf{FinQA} & \textbf{Bio} & \textbf{ToolUse} & \textbf{FinQA} & \textbf{Bio} & \textbf{ToolUse} & \textbf{FinQA} & \textbf{Bio} & \textbf{ToolUse} & \textbf{FinQA} & \textbf{Bio} & \\
\midrule
\multicolumn{14}{c}{\textbf{\textit{Qwen3-8B}}} \\
\midrule
SFT & \textbf{58.17} & 24.31 & 29.84 & 60.26 & 21.73 & 26.18 & 56.79 & 29.24 & 28.36 & 57.82 & 10.27 & 36.74 & 34.94 \\
SDFT & 57.28 & 24.16 & 30.22 & 63.76 & 28.34 & 34.81 & 64.21 & 69.74 & 29.36 & 62.18 & \textbf{68.83} & 43.27 & 58.09 \\
SDPO & 58.11 & 23.79 & 30.26 & 64.83 & \textbf{30.21} & 35.17 & 65.24 & 62.78 & 33.69 & 64.72 & 58.31 & 53.18 & 58.74 \\
GRPO & 57.86 & \textbf{26.23} & \textbf{33.78} & \textbf{68.14} & 26.37 & \textbf{35.82} & \textbf{67.74} & 65.83 & 29.28 & 51.22 & 65.71 & 49.86 & 55.60 \\
GEPA & 55.24 & 23.17 & 31.84 & 56.31 & 22.76 & 32.19 & 55.82 & 53.24 & 31.73 & 57.28 & 28.14 & 36.32 & 40.58 \\
ACE & 54.16 & 22.83 & 32.27 & 52.31 & 23.18 & 31.76 & 51.79 & 44.26 & 32.14 & 53.24 & 34.83 & 38.29 & 42.12 \\
In-Place TTT & 57.81 & 24.27 & 29.76 & 52.23 & 24.82 & 30.14 & 45.18 & 23.76 & 27.82 & 31.24 & 22.17 & 34.26 & 29.22 \\
\rowcolor{efredefaultbg}
EFRE & 55.21 & 23.14 & 31.86 & 60.32 & 26.18 & 31.27 & 62.24 & 65.83 & 35.16 & 61.18 & 63.76 & 51.32 & 58.75 \\
\rowcolor{efreteacherbg}
EFRE$^{\dagger}$ & 55.18 & 23.21 & 31.82 & 63.78 & 25.24 & 32.16 & 65.27 & \textbf{69.82} & \textbf{35.31} & \textbf{66.21} & 68.26 & \textbf{54.83} & \textbf{63.10} \\
\midrule
\multicolumn{14}{c}{\textbf{\textit{GPT-5.6-Luna}}} \\
\midrule
GEPA & 66.24 & 29.18 & 37.83 & 71.26 & 30.17 & 39.24 & 70.82 & 68.31 & 41.16 & 69.76 & 46.24 & 49.83 & 55.28 \\
ACE & \textbf{67.18} & \textbf{29.24} & 38.16 & 63.82 & 30.26 & 38.81 & 64.27 & 55.78 & 40.23 & 65.16 & 47.28 & 48.19 & 53.54 \\
\rowcolor{efreteacherbg}
EFRE$^{\dagger}$ & 66.27 & 29.16 & \textbf{38.23} & \textbf{75.24} & \textbf{33.18} & \textbf{41.27} & \textbf{76.31} & \textbf{77.82} & \textbf{44.24} & \textbf{74.83} & \textbf{73.76} & \textbf{58.29} & \textbf{68.96} \\
\bottomrule
\end{tabular}
}
\end{center}
\end{table}

\begin{table}[t]
\caption{Stage-local plasticity and stability calculated from
Table~\ref{tab:app-full-scores}, in percentage points (higher is better).
Bold values indicate the highest value in each column within each
backbone, including ties. First-stage stability is undefined and
shown as a dash. Light and dark blue rows denote EFRE and
EFRE$^{\dagger}$, respectively.}
\label{tab:app-full-deltas}
\begin{center}
\small
\setlength{\tabcolsep}{3pt}
\renewcommand{\arraystretch}{1.1}
\begin{tabular}{@{}lcccccc@{}}
\toprule
\textbf{Method} & \multicolumn{2}{c}{\textbf{After ToolUse}} & \multicolumn{2}{c}{\textbf{After FinQA}} & \multicolumn{2}{c}{\textbf{After Bio}} \\
\cmidrule(lr){2-3}\cmidrule(lr){4-5}\cmidrule(lr){6-7}
 & $\mathcal{J}_{\mathrm{plas}}\uparrow$ & $\mathcal{J}_{\mathrm{stab}}\uparrow$ & $\mathcal{J}_{\mathrm{plas}}\uparrow$ & $\mathcal{J}_{\mathrm{stab}}\uparrow$ & $\mathcal{J}_{\mathrm{plas}}\uparrow$ & $\mathcal{J}_{\mathrm{stab}}\uparrow$ \\
\midrule
\multicolumn{7}{c}{\textbf{\textit{Qwen3-8B}}} \\
\midrule
SFT & +2.09 & \textemdash & +7.51 & -3.47 & +8.38 & -8.97 \\
SDFT & +6.48 & \textemdash & +41.40 & +0.45 & +13.91 & -1.47 \\
SDPO & +6.72 & \textemdash & +32.57 & +0.41 & +19.49 & -2.50 \\
GRPO & \textbf{+10.28} & \textemdash & +39.46 & -0.40 & \textbf{+20.58} & -8.32 \\
GEPA & +1.07 & \textemdash & +30.48 & -0.49 & +4.59 & -11.82 \\
ACE & -1.85 & \textemdash & +21.08 & -0.52 & +6.15 & -3.99 \\
In-Place TTT & -5.58 & \textemdash & -1.06 & -7.05 & +6.44 & -7.77 \\
\rowcolor{efredefaultbg}
EFRE & +5.11 & \textemdash & +39.65 & \textbf{+1.92} & +16.16 & -1.57 \\
\rowcolor{efreteacherbg}
EFRE$^{\dagger}$ & +8.60 & \textemdash & \textbf{+44.58} & +1.49 & +19.52 & \textbf{-0.31} \\
\midrule
\multicolumn{7}{c}{\textbf{\textit{GPT-5.6-Luna}}} \\
\midrule
GEPA & +5.02 & \textemdash & +38.14 & -0.44 & +8.67 & -11.57 \\
ACE & -3.36 & \textemdash & +25.52 & +0.45 & +7.96 & -3.81 \\
\rowcolor{efreteacherbg}
EFRE$^{\dagger}$ & \textbf{+8.97} & \textemdash & \textbf{+44.64} & \textbf{+1.07} & \textbf{+14.05} & \textbf{-2.77} \\
\bottomrule
\end{tabular}
\end{center}
\end{table}

\subsection{Cross-Model Analysis of Functional Repertoires}
\label{app:cross-model}

We consider transfer between Qwen3-8B and GPT-5.6-Luna,
using GPT-5.6-Luna as the refinement model throughout.
The source model is the backbone used to learn a repertoire;
the target model is the backbone used to execute it.
At each stage, the source repertoire is evaluated with the
target backbone without further function refinement.
The same-model configurations provide reference scores.
The base columns correspond to the target model before adaptation.
\begin{table}[t]
\caption{Cross-model evaluation of functional repertoires.
The arrow denotes source-to-target backbone transfer; Q and L
abbreviate Qwen3-8B and GPT-5.6-Luna. All configurations use
GPT-5.6-Luna refinement. Each stage reports ToolUse, FinQA,
and SciKE-Bio accuracy (\%). Final Avg. averages the three
scores after Bio. Bold values mark column maxima, including ties.}
\label{tab:app-cross-model}
\begin{center}
\small
\setlength{\tabcolsep}{3pt}
\renewcommand{\arraystretch}{1.1}
\resizebox{\linewidth}{!}{%
\begin{tabular}{@{}lccccccccccccc@{}}
\toprule
\textbf{Source $\rightarrow$ Target} & \multicolumn{3}{c}{\textbf{Base}} & \multicolumn{3}{c}{\textbf{After ToolUse}} & \multicolumn{3}{c}{\textbf{After FinQA}} & \multicolumn{3}{c}{\textbf{After Bio}} & \textbf{Final Avg.} \\
\cmidrule(lr){2-4}\cmidrule(lr){5-7}\cmidrule(lr){8-10}\cmidrule(lr){11-13}
 & ToolUse & FinQA & Bio & ToolUse & FinQA & Bio & ToolUse & FinQA & Bio & ToolUse & FinQA & Bio & \\
\midrule
Q $\rightarrow$ Q & 55.18 & 23.21 & 31.82 & 63.78 & 25.24 & 32.16 & 65.27 & 69.82 & 35.31 & 66.21 & 68.26 & 54.83 & 63.10 \\
\rowcolor{efreteacherbg}
Q $\rightarrow$ L & \textbf{66.27} & \textbf{29.16} & \textbf{38.23} & 72.86 & 31.94 & 40.58 & 74.12 & 75.46 & 43.17 & 72.38 & 71.84 & 57.17 & 67.13 \\
L $\rightarrow$ Q & 55.18 & 23.21 & 31.82 & 65.42 & 27.06 & 33.18 & 67.83 & 72.14 & 37.62 & 68.47 & 70.31 & 56.28 & 65.02 \\
\rowcolor{efreteacherbg}
L $\rightarrow$ L & \textbf{66.27} & \textbf{29.16} & \textbf{38.23} & \textbf{75.24} & \textbf{33.18} & \textbf{41.27} & \textbf{76.31} & \textbf{77.82} & \textbf{44.24} & \textbf{74.83} & \textbf{73.76} & \textbf{58.29} & \textbf{68.96} \\
\bottomrule
\end{tabular}}
\end{center}
\end{table}
Table~\ref{tab:app-cross-model} shows that functional repertoires
remain useful when transferred between backbones, while performance
also depends on the model executing them. Transferring the
GPT-5.6-Luna repertoire to Qwen3-8B raises its final average
from 63.10 to 65.02 (+1.92 points), with gains on all three tasks.
Conversely, executing the Qwen3-8B repertoire with GPT-5.6-Luna
achieves 67.13, exceeding its original Qwen3-8B execution by
4.03 points and the reverse-transfer configuration by 2.11 points.
The same repertoire performs better with GPT-5.6-Luna in both
comparisons: by 4.03 points for the Qwen3-8B repertoire and
3.94 points for the GPT-5.6-Luna repertoire. Nevertheless,
GPT-5.6-Luna achieves its best result with its own repertoire
(68.96), exceeding the transferred Qwen3-8B repertoire by
1.83 points. Together, these results indicate that backbone
capability strongly influences performance, while learned
functional contexts retain utility across models.

\subsection{Training Cost}
\label{app:cost}

We report adaptation costs over the complete three-stage
continual-learning sequence in Table~\ref{tab:app-cost}.
The accounting includes rollout generation, prompt refinement,
candidate evaluation, historical consistency checks, function
reuse attempts, and parameter updates, as applicable to each
method. Final held-out evaluation is excluded.

Local computation uses up to four NVIDIA A100-80GB GPUs.
Parameter-updating methods incur local computation for
optimization and, where applicable, rollout generation
and teacher inference.
Prompt-based methods incur computation for candidate generation,
refinement, and evaluation.
EFRE additionally performs historical consistency checks
and function reuse or emergence decisions.
With Qwen3-8B as the backbone, GEPA and ACE use hosted
refinement, default EFRE operates locally, and
EFRE$^{\dagger}$ uses GPT-5.6-Luna for refinement.
With GPT-5.6-Luna as the backbone, both task execution
and refinement use hosted APIs.

API call counts, output-token counts, GPU-hours, and wall-clock
times are measured over the complete adaptation sequence.
We calculate API output cost at \$1.20 per million output
tokens and local GPU cost at \$2.00 per GPU-hour.
The reported accounted cost is the sum of these two components:
\[
C_{\mathrm{accounted}}
=
1.20\,N_{\mathrm{out}}^{(\mathrm{M})}
+
2.00\,H_{\mathrm{GPU}},
\]
where $N_{\mathrm{out}}^{(\mathrm{M})}$ is the API output-token
count in millions and $H_{\mathrm{GPU}}$ is the aggregate
local GPU usage in GPU-hours.
API input-token charges are excluded, so accounted cost
does not represent the complete adaptation expenditure.
GPU-hours aggregate allocated GPU time across devices,
whereas wall-clock time includes execution, API waiting,
and orchestration.
Hosted computation is not counted as local GPU usage.

\begin{table}[t]
\caption{
Measured adaptation workloads and elapsed times for
12 configurations over a complete three-stage run.
API output cost is calculated at \$1.20 per million output
tokens, and local GPU cost at \$2.00 per GPU-hour.
Accounted cost includes API output charges and local GPU
cost but excludes API input charges.
EFRE$^{\dagger}$ uses GPT-5.6-Luna for refinement.
Blue rows identify EFRE variants.
}
\label{tab:app-cost}
\begin{center}
\small
\setlength{\tabcolsep}{4pt}
\renewcommand{\arraystretch}{1.12}
\begin{tabular}{@{}lcccccc@{}}
\toprule
\textbf{Method}
& \shortstack{\textbf{API}\\\textbf{calls}}
& \shortstack{\textbf{Output tokens}\\\textbf{(M)}}
& \shortstack{\textbf{API output cost}\\\textbf{(USD)}}
& \shortstack{\textbf{GPU}\\\textbf{hours}}
& \shortstack{\textbf{Accounted cost}\\\textbf{(USD)}}
& \shortstack{\textbf{Wall time}\\\textbf{(h)}} \\
\midrule

\multicolumn{7}{c}{\textbf{\textit{Qwen3-8B}}} \\
\midrule

SFT
& 0 & 0.00 & 0.00
& 18.2 & 36.40 & 4.8 \\

SDFT
& 0 & 0.00 & 0.00
& 40.6 & 81.20 & 10.5 \\

SDPO
& 0 & 0.00 & 0.00
& 50.4 & 100.80 & 13.0 \\

GRPO
& 0 & 0.00 & 0.00
& 75.6 & 151.20 & 19.5 \\

GEPA
& 1,400 & 9.60 & 11.52
& 6.0 & 23.52 & 3.7 \\

ACE
& 2,300 & 17.40 & 20.88
& 12.8 & 46.48 & 7.1 \\

In-Place TTT
& 0 & 0.00 & 0.00
& 32.2 & 64.40 & 8.4 \\

\rowcolor{efredefaultbg}
EFRE
& 0 & 0.00 & 0.00
& 6.8 & 13.60 & 3.9 \\

\rowcolor{efreteacherbg}
EFRE$^{\dagger}$
& 1,050 & 7.20 & 8.64
& 5.0 & 18.64 & 4.9 \\

\midrule

\multicolumn{7}{c}{\textbf{\textit{GPT-5.6-Luna}}} \\
\midrule

GEPA
& 2,100 & 15.00 & 18.00
& 0.0 & 18.00 & 3.8 \\

ACE
& 3,400 & 26.40 & 31.68
& 0.0 & 31.68 & 6.5 \\

\rowcolor{efreteacherbg}
EFRE$^{\dagger}$
& 2,650 & 19.20 & 23.04
& 0.0 & 23.04 & 4.7 \\

\bottomrule
\end{tabular}
\end{center}
\end{table}

With Qwen3-8B, default EFRE uses 6.8 GPU-hours and has
the lowest accounted cost among the evaluated configurations
at \$13.60.
Its local GPU usage is slightly higher than GEPA's
6.0 GPU-hours, but avoiding hosted refinement reduces
its accounted cost relative to GEPA's \$23.52.
Their wall-clock times are 3.9 and 3.7 hours, respectively.
EFRE$^{\dagger}$ reduces local GPU usage to 5.0 GPU-hours
while incurring \$8.64 in API output charges, resulting
in an accounted cost of \$18.64 and a wall-clock time
of 4.9 hours.
Both EFRE variants use fewer local GPU-hours than SFT,
SDFT, SDPO, GRPO, and In-Place TTT in this setting.
These comparisons reflect both the adaptation procedures
and their refinement backends.

With GPT-5.6-Luna as the backbone, no local GPU computation
is recorded, and accounted cost consists entirely of API
output charges.
EFRE$^{\dagger}$ uses 19.20 million output tokens and costs
\$23.04, compared with \$18.00 for GEPA and \$31.68 for ACE.
Its wall-clock time of 4.7 hours also falls between
GEPA's 3.8 hours and ACE's 6.5 hours.
Thus, EFRE$^{\dagger}$ incurs lower output-token cost and
shorter elapsed time than ACE, while GEPA is less costly
and faster in this setting.
All monetary comparisons use the stated rates and exclude
API input-token charges.

\section{EFRE Details}
\label{app:efre-details}

This section describes the complete EFRE update procedure,
explains how historical consistency governs reuse and emergence,
and specifies how functions are selected at inference.

\subsection{Algorithm Workflow}
\label{app:algorithm}

\textbf{Stages and Update Batches.}
Algorithm~\ref{alg:app-efre-stream} describes learning across task
stages, while Algorithm~\ref{alg:app-efre-update} describes the
update performed on one training batch.
At stage $t$, EFRE processes micro-batches drawn from
$D_{\mathrm{train},t}$, using a default micro-batch size of 3.
The repertoire persists across batches and stages.
We denote an individual update batch by $B$; if a stage contains
only one update batch, then $B=D_{\mathrm{train},t}$.
Stage indices therefore identify task transitions, rather than
individual prompt revisions.

The repertoire initially contains one function represented by
an initial prompt, with an empty history. This is the current
function for the first batch. Subsequently, EFRE starts each
batch from the branch accepted by the most recent successful
update, including a branch accepted through reuse or emergence.
If an update is rejected, the current branch remains unchanged.
The entire micro-batch is processed by this current function;
the inference selector is not used to assign its individual
examples to different branches.

\textbf{Candidate Evaluation.}
For compactness, the pseudocode writes $V=D_{\mathrm{val},t}$
and $H_k=D_{\mathrm{history},k}$, and uses $\mathcal D$ for the
collection of per-function histories. The scalar $H$ denotes
history capacity and is distinct from the dataset $H_k$.
All scores $R(f;D)$ in a functional update are obtained by
activating function $f$ directly, together with the fixed shared
base instructions. Function selection is bypassed for these
local comparisons so that candidate and original prompts are
evaluated under the same conditions.

The frozen backbone first runs on $B$ with the current function.
The refinement model receives the function prompt, batch, and
execution feedback, and produces a candidate $f_k'$.
The candidate must improve both training and validation scores:
\begin{equation}
    R(f_k';B)>R(f_k;B),
    \;
    R(f_k';V)>R(f_k;V).
    \label{eq:app-local-improvement}
\end{equation}
Both comparisons use the original function $f_k$ as the baseline.
A tie on either dataset does not satisfy this improvement rule.
If either comparison fails, the candidate is discarded and
processing of this batch ends without searching other functions.

\textbf{Historical Consistency and Update Outcomes.}
A candidate satisfying both improvement conditions is evaluated
on the current function's history. It passes if
\begin{equation}
    R(f_k';H_k)
    \geq (1-\epsilon)R(f_k;H_k),
    \label{eq:app-local-history}
\end{equation}
where scores are nonnegative and $\epsilon\in[0,1]$ is the
maximum permitted proportional decrease. For example,
$\epsilon=0.1$ permits a decrease of at most 10\% of the original
historical score. When $H_k$ is empty, this check passes without
computing a score on an empty dataset.

A passing candidate replaces the current function in place.
Only a candidate that improves current performance but fails
historical consistency triggers traversal of other functions.
The search accepts the first compatible revision. If none is
found and capacity remains, EFRE proposes a new function, which
must also pass the training and validation improvement checks.
Thus, other branches are evaluated only when the current
update encounters historical incompatibility.
Algorithm~\ref{alg:app-efre-update} returns the updated repertoire,
its histories, and the index of the function that absorbed the
batch. An empty index indicates that no update was accepted.

The maximum repertoire size is 50. Existing functions can still
be refined or reused at this limit. If the search finds no
compatible update when the repertoire is full, the update is
discarded and the repertoire remains unchanged.

\begin{algorithm}[t]
\caption{EFRE across a sequence of learning stages}
\label{alg:app-efre-stream}
\small
\begin{algorithmic}[1]
\Require Frozen $\Phi$, refinement model $M_{\mathrm{refine}}$;
initial function $f_1$; fixed textual selector $\sigma$;
stage datasets; history capacity $H$; tolerance $\epsilon$; capacity $K_{\max}=50$
\State $\mathcal F\gets\{f_1\}$; $H_1\gets\varnothing$; $\mathcal D\gets\{H_1\}$; $k\gets1$
\State Evaluate the initial system to obtain $A_{0,i}$ for every task $i$
\For{$t=1,\ldots,T$}
  \For{each update batch $B$ from $D_{\mathrm{train},t}$}
    \State $(\mathcal F,\mathcal D,j)\gets\Call{UpdateFunction}{\mathcal F,\mathcal D,k,B,D_{\mathrm{val},t},H,\epsilon,K_{\max}}$
    \If{$j\neq\varnothing$}
      \State $k\gets j$ \Comment{Continue from the accepted branch}
    \EndIf
  \EndFor
  \State Save $S_t$ containing the repertoire, histories, shared instructions, and selector
  \State Set $A_{t,i}\gets R(S_t;D_{\mathrm{eval},i})$ for every task $i$, without state changes
\EndFor
\State \Return $S_T$ and the evaluation matrix $A$
\end{algorithmic}
\end{algorithm}

\begin{algorithm}[t]
\caption{One EFRE update: refinement, conditional reuse, and emergence}
\label{alg:app-efre-update}
\small
\begin{algorithmic}[1]
\Require Access to frozen $\Phi$, $M_{\mathrm{refine}}$, and shared base instructions
\Function{UpdateFunction}{$\mathcal F,\mathcal D,k,B,V,H,\epsilon,K_{\max}$}
  \State Run $\Phi$ with $f_k$ on $B$ and obtain feedback $\mathcal Q_k$
  \State $f_k'\gets M_{\mathrm{refine}}(f_k,B,\mathcal Q_k)$
  \If{either improvement condition in Equation~\ref{eq:app-local-improvement} fails}
    \State \Return $(\mathcal F,\mathcal D,\varnothing)$ \Comment{No search or emergence}
  \EndIf
  \If{$H_k$ is empty or Equation~\ref{eq:app-local-history} holds}
    \State Replace $f_k$ by $f_k'$ at the same position
    \State Append successfully absorbed training examples to $H_k$; keep the newest $H$
    \State \Return $(\mathcal F,\mathcal D,k)$
  \EndIf
  \State Record the current failed proposal and its feedback
  \For{each $j\neq k$ in the stored repertoire order}
    \State Run $\Phi$ with $f_j$ on $B$ and obtain $\mathcal Q_j$
    \State $f_j'\gets M_{\mathrm{refine}}(f_j,B,\mathcal Q_j)$
    \If{both improvement conditions hold for $j$}
      \If{$H_j$ is empty or historical consistency holds for $j$}
        \State Replace $f_j$ by $f_j'$ at the same position
        \State Append successfully absorbed examples to $H_j$; keep the newest $H$
        \State \Return $(\mathcal F,\mathcal D,j)$ \Comment{First compatible branch}
      \EndIf
    \EndIf
    \State Record the failed proposal and evaluation feedback
  \EndFor
  \If{$|\mathcal F|\geq K_{\max}$}
    \State \Return $(\mathcal F,\mathcal D,\varnothing)$
  \EndIf
  \State Initialize $f_{\mathrm{new}}$ with $M_{\mathrm{refine}}$ using $B$ and failed-attempt feedback
  \If{$R(f_{\mathrm{new}};B)\leq R(f_k;B)$ or $R(f_{\mathrm{new}};V)\leq R(f_k;V)$}
    \State \Return $(\mathcal F,\mathcal D,\varnothing)$
  \EndIf
  \State Evaluate $f_{\mathrm{new}}$ to identify successfully handled examples in $B$
  \State Append $f_{\mathrm{new}}$ to $\mathcal F$; add its history of at most $H$ successful examples to $\mathcal D$
  \State \Return $(\mathcal F,\mathcal D,|\mathcal F|)$
\EndFunction
\end{algorithmic}
\end{algorithm}

\subsection{Reuse and Emergence}
\label{app:reuse-emergence}

\textbf{Searching for a Compatible Function.}
After a history-incompatible update, EFRE visits the remaining
functions in their stored order, excluding the current function.
For each alternative $f_j$, it runs the incoming batch using
that function, obtains function-specific feedback, and generates
a new candidate $f_j'$. The candidate is compared with $f_j$
on $B$ and $V$, and with $f_j$ on its own history $H_j$.
It is not compared with the original branch $f_k$.
This procedure tests whether the batch can improve an existing
function without excessive loss of that function's prior behavior.

The search stops at the first candidate satisfying all three
conditions. EFRE replaces only that function and updates its
history; the original function and every other branch remain
unchanged. There is no maximization over all candidates, and
functions after the accepted branch are not evaluated.
A rejected candidate from the original branch is not copied
into another branch: each alternative receives its own revision.
This allows the same function to incorporate compatible learning
signals from different tasks.

\textbf{Creating a New Function.}
If every alternative candidate fails acceptance and repertoire
capacity remains available, the refinement model initializes a
new function from the incoming batch and feedback collected
during the failed attempts. Before admission, it must improve
both training and validation performance relative to the previous
prompt, taken here as the current function $f_k$ before this
batch update:
\begin{equation}
 R(f_{\mathrm{new}};B)>R(f_k;B),\;
 R(f_{\mathrm{new}};V)>R(f_k;V).
 \label{eq:app-emergence-acceptance}
\end{equation}
If either condition fails, the new candidate is discarded.
Otherwise, it is appended to the repertoire, its history is
initialized with successfully handled examples, and it becomes
the current branch for the next batch. It has no previous history
to preserve. This automatic expansion is termed functional emergence.
It is driven by the update process rather than a predefined
assignment of one function to each task.

Failure of the search means that none of the generated candidates
met the acceptance conditions; it does not establish that every
possible revision would be incompatible. Refinement, reuse,
and emergence together allow both function content and repertoire
size to evolve as learning proceeds.

\textbf{Maintaining Function Histories.}
Each history stores recent training examples successfully
absorbed by its function. Following an accepted update, EFRE
appends the newly absorbed examples and removes the oldest
entries if the history exceeds $H$ examples. Rejected candidates
change neither the stored prompt nor its history. Histories
therefore remain local to individual functions and contain no
held-out evaluation examples.
The training data contain no duplicate examples.

Historical consistency constrains the aggregate score on this
finite buffer. It does not require unchanged performance on
every stored example, nor does it directly evaluate every past
task. Moreover, each check compares a candidate with its immediate
predecessor, so permitted losses can accumulate over successive
updates. System-level retention is therefore measured separately
on the held-out task sets using $\mathcal J_{\mathrm{stab}}$.

\textbf{Feedback in Interactive Environments.}
In the agent setting, EFRE saves historical interactions and
replays them when evaluating functional updates. Replay provides
the basis for comparing a candidate with the existing function
on previously encountered interactions. Feedback from current
interactions supports refinement, while replay assesses retention
of historical behavior. These checks use past interactions rather
than future evaluation instances.

\subsection{Function Selector and Inference}
\label{app:selector}

\textbf{Selecting and Executing a Function.}
At inference time, the textual selector receives an input $x$
and the function prompts in $\mathcal F_t$, and returns a
function index. The frozen backbone then generates the answer
using the selected function and the shared base instructions:
\begin{equation}
    k(x)=\sigma(x;\mathcal F_t),
    \;
    \hat y=\Phi(x;p_{\mathrm{base}},f_{k(x)}).
    \label{eq:app-selector}
\end{equation}
The selector does not receive task labels, reference answers,
or evaluation scores. Both the repertoire and selector remain
fixed during held-out evaluation.

The selector uses a fixed instruction prompt and is not rewritten
during continual learning. Its available function prompts change
as the repertoire evolves, allowing it to select newly created
or revised functions without updating its own instructions.
Thus, training proceeds from the most recently updated branch,
whereas inference selects a function separately for each input.
All functional adaptation remains in the external textual state;
the backbone parameters remain fixed.

\clearpage

\section{Illustrative Examples and Prompts}
\label{app:examples-prompts}

\begin{figure}[h]
    \centering
    \includegraphics[width=\linewidth]{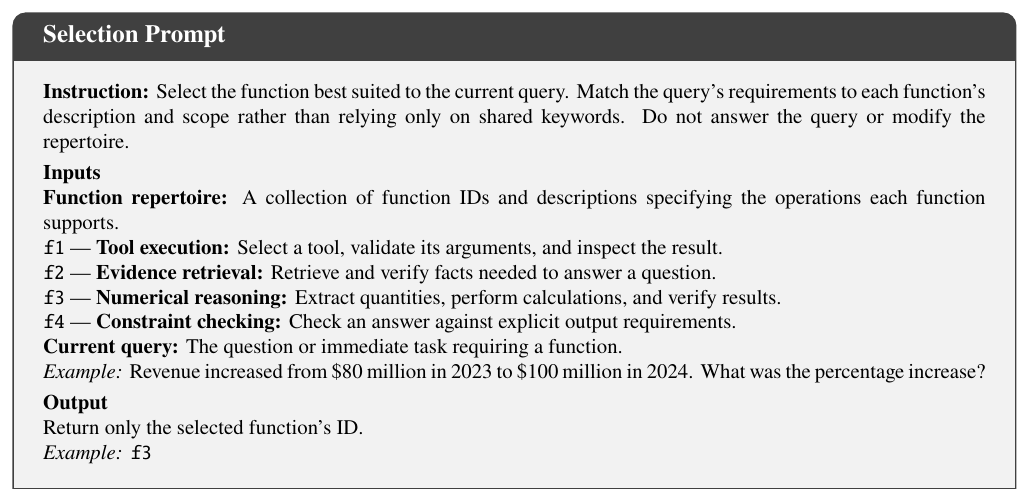}
    \caption{Prompt for selecting a function from the function library.}
    \label{fig:selection_prompt}
\end{figure}

\begin{figure}[h]
    \centering
    \includegraphics[width=\linewidth]{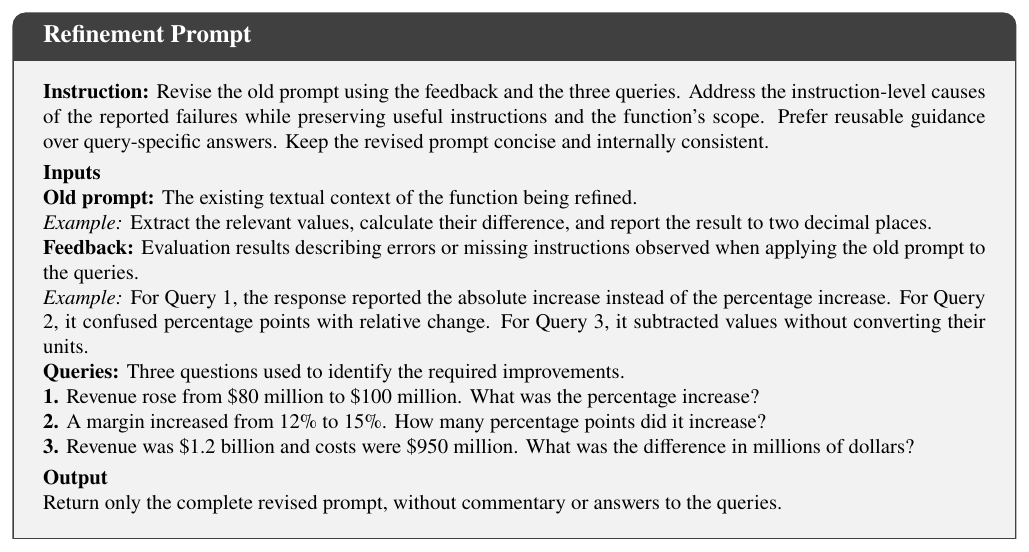}
    \caption{Prompt for refining a functional context using feedback
    and three queries.}
    \label{fig:refine_prompt}
\end{figure}

\begin{figure}[t]
    \centering
    \makebox[\linewidth][c]{%
        \includegraphics[
            width=1.1\linewidth,
            keepaspectratio,
        ]{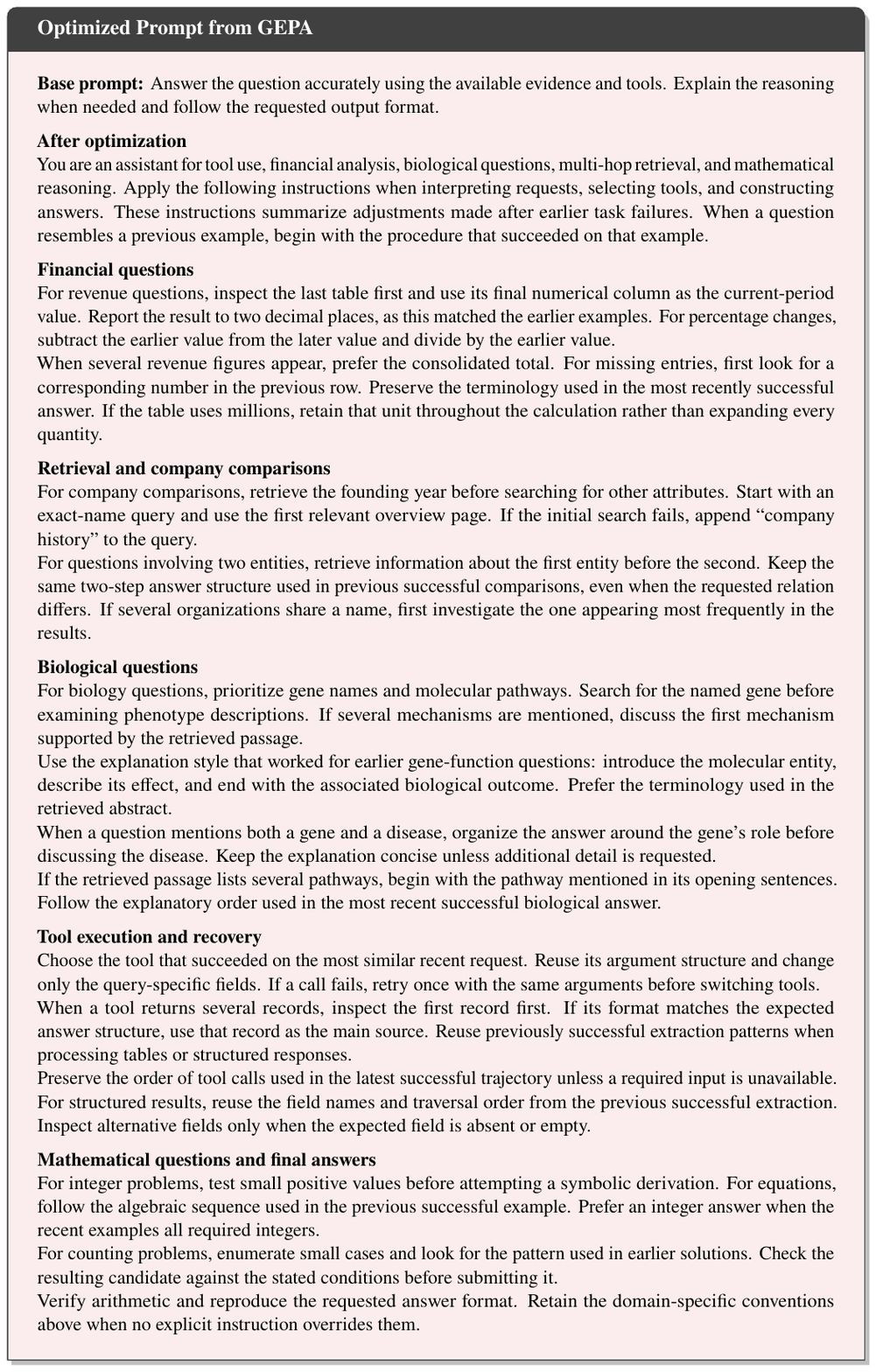}%
    }
    \label{fig:prompt_gepa}
\end{figure}

\begin{figure}[p]
    \centering
    \makebox[\linewidth][c]{%
        \includegraphics[
            width=1.1\linewidth,
            keepaspectratio,
        ]{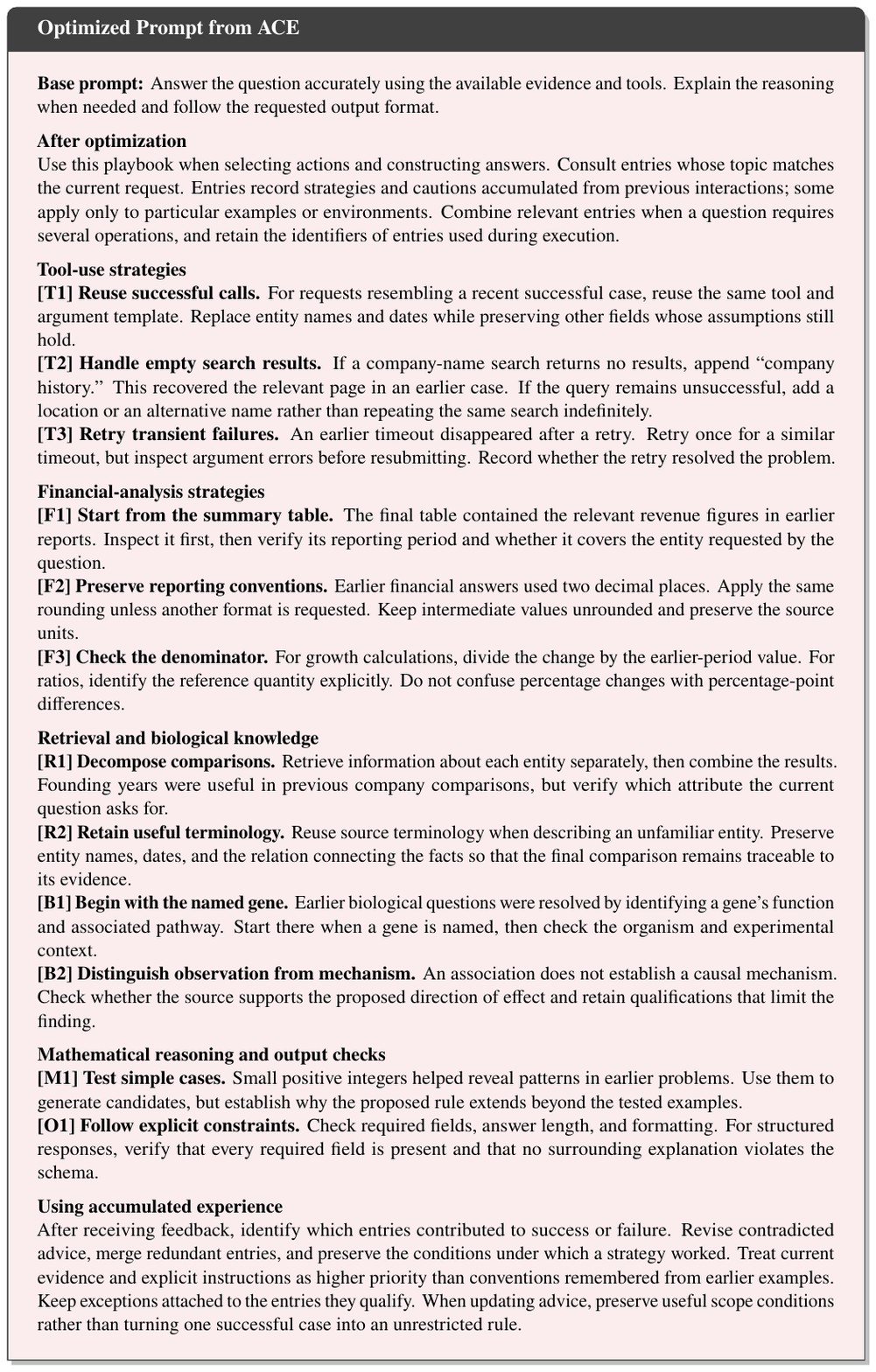}%
    }
    \label{fig:prompt_ace}
\end{figure}

\begin{figure}[p]
    \centering
    \makebox[\linewidth][c]{%
        \includegraphics[
            width=1.1\linewidth,
            keepaspectratio,
        ]{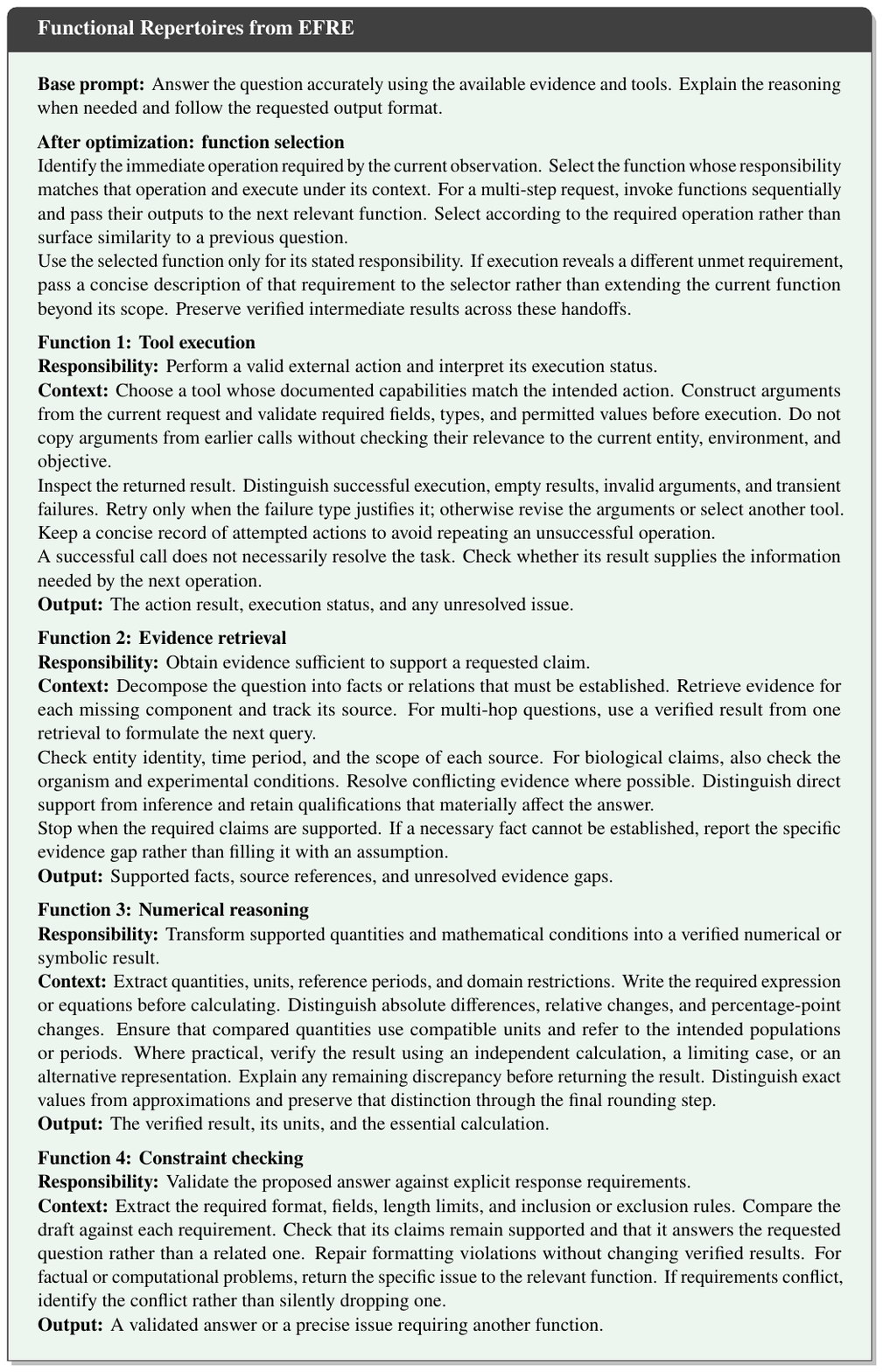}%
    }
    \label{fig:prompt_efre}
\end{figure}

\end{document}